%% file: main.tex
\documentclass{article}
\usepackage{graphicx}
\usepackage{enumitem}
\usepackage{tcolorbox}
\usepackage{listings}
\usepackage{booktabs}
\usepackage{multirow}
\usepackage{graphicx}
\usepackage[T1]{fontenc}

\usepackage{wrapfig}
\usepackage{graphicx}

\usepackage{newtxtext,newtxmath} 

\usepackage{xcolor}
\usepackage[preprint]{corl_2026}

\newcommand{\ours}{AssemLM}

\title{\ours: A Spatial Reasoning Multimodal Large Language Model for Robotic Assembly}

\author{
  Zhi Jing\textsuperscript{1,2} \quad
  Jinbin Qiao\textsuperscript{2,3} \quad
  Ouyang Lu\textsuperscript{2,4} \quad
  Jicong Ao\textsuperscript{2}\\
  \textbf{Shuang Qiu\textsuperscript{6}\quad 
  Huazhe Xu\textsuperscript{5}\quad 
  Yu-Gang Jiang\textsuperscript{1*}\quad 
  Chenjia Bai\textsuperscript{2*}}\\
  \\
  \textsuperscript{1}Fudan University\textsuperscript{\dag} \quad
  \textsuperscript{2}Institute of Artificial Intelligence (TeleAI), China Telecom\textsuperscript{\dag}
  \textsuperscript{3}Tianjin University\quad \\
  \textsuperscript{4}Northwestern Polytechnical University\quad
  \textsuperscript{5}Tsinghua University\quad
  \textsuperscript{6}City University of Hong Kong\\
  \textsuperscript{*}Corresponding author \quad
  \textsuperscript{\dag}Equally leading organizations\\[0.3em]
  \\
  \url{https://assemlmhome.github.io/}
}

\begin{document}
\maketitle

\input{contents/fig_main}
\input{contents/00_abstract}


\input{contents/01_into}
\input{contents/02_related_work}
\input{contents/03_method}
\input{contents/04_experiments}
\input{contents/05_conclusion}

\bibliography{main}

\clearpage
\newpage
\pagebreak
\input{contents/99_appendix}

\end{document}

%% file: contents/fig_main.tex
\begin{figure}[h!]
    \vspace{-.3in}
    \centering
    \includegraphics[width=1\linewidth]{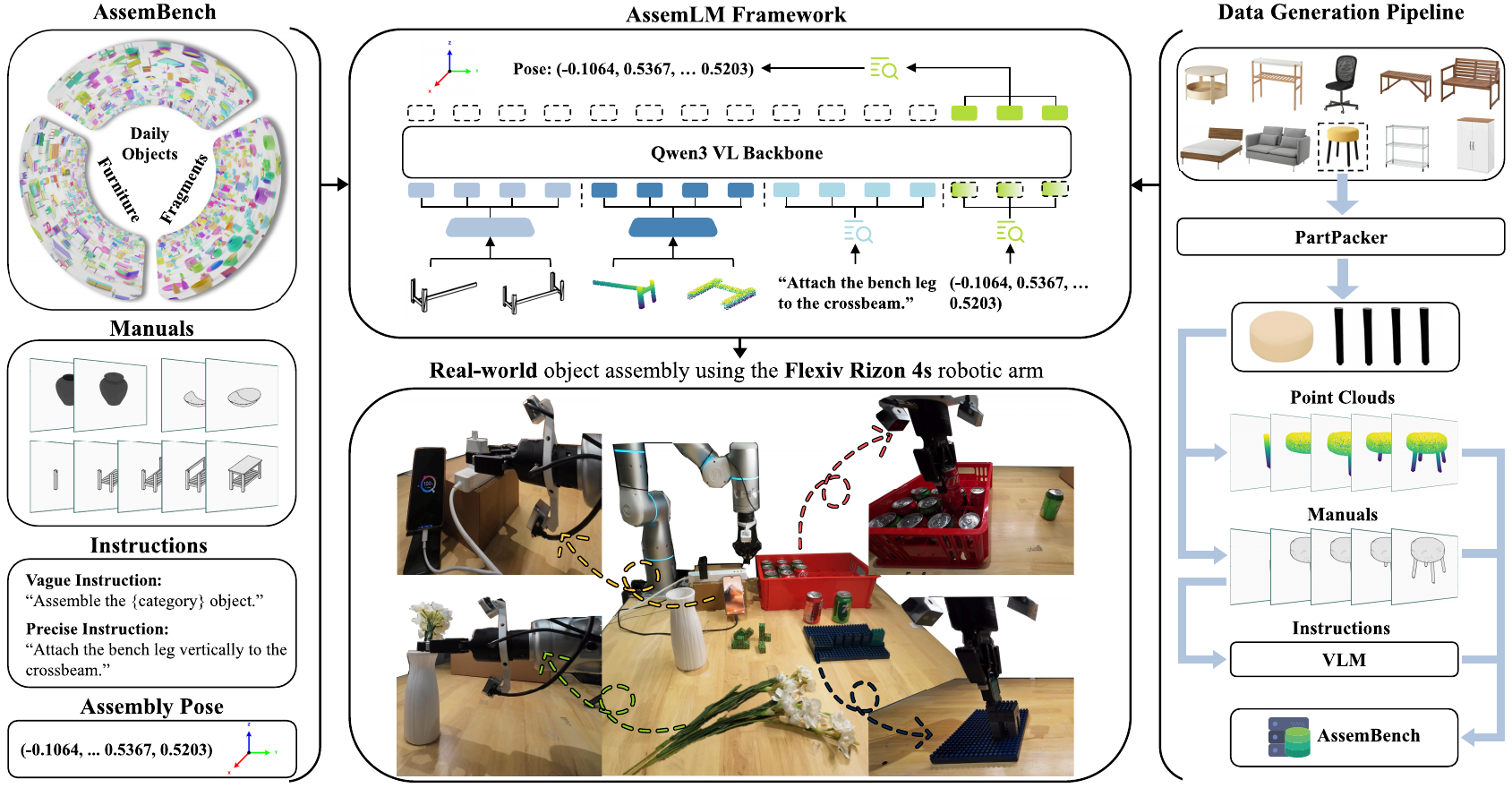}
    \caption{
    \textbf{AssemLM and AssemBench: Scaling Spatial Reasoning for Robotic Assembly.}
    (Left \& Right) We introduce \textit{AssemBench}, a large-scale multimodal dataset and its generation pipeline, providing rich object categories, manuals, text instructions, and 6D pose annotations. 
    (Middle) The proposed \textit{AssemLM} architecture (top) processes these multimodal inputs by coupling $SO(3)$-equivariant geometric features with a Qwen3-VL backbone to predict 6D assembly poses, facilitating real-world execution of fine-grained tasks on a Flexiv Rizon 4s arm (bottom).
    }
    \label{fig:main}
    \vspace{-2.5mm}
\end{figure}

%% file: contents/00_abstract.tex
\begin{abstract}
Spatial reasoning is a fundamental capability for embodied intelligence, especially for fine-grained manipulation tasks such as robotic assembly.
Recent methods based on vision-language models (VLMs) largely rely on coarse 2D perception and struggle to perform accurate reasoning over complex 3D geometry.
To address this limitation, we propose \ours, a spatial multimodal large language model for robotic assembly that integrates assembly manuals, point clouds, and textual instructions to predict task-critical 6D assembly poses with explicit geometric understanding. 
To bridge raw 3D perception and high-level linguistic reasoning, \ours~employs a specialized point cloud encoder to extract fine-grained geometric and rotational features for accurate 3D spatial reasoning in assembly tasks.
In addition, we introduce AssemBench, a large-scale benchmark for assembly-oriented spatial reasoning with over 900K multimodal samples and precise 6D pose annotations, extending evaluation from 2D grounding to full 3D geometric inference.
Extensive experiments and real-robot evaluations demonstrate that \ours~achieves state-of-the-art 6D pose reasoning performance and effectively supports fine-grained, multi-step assembly tasks in real-world settings. Code, models, and the AssemBench dataset will be made publicly available.
\end{abstract}

%% file: contents/01_into.tex
\vspace{-0.03in}
\section{Introduction}
\vspace{-0.03in}
From assembling furniture~\cite{ben2021ikea,heo2025furniturebench,liang2026a3d,tie2025manual2skill++,tie2025manual2skill} to repairing appliances~\cite{gao2025realappliance,long2025checkmanual} and reconstructing fragmented objects~\cite{islam2025m3rf,sellan2022breaking,shen2025biassemble,wu2023leveraging}, assembly processes are ubiquitous in daily life.
Robotic assembly assistance can reduce human effort and improve accessibility for non-expert users.
However, these tasks remain highly challenging for embodied intelligence, requiring fine-grained geometric reasoning over task-critical 6D poses, joint understanding of multimodal inputs including text, images, and 3D data, and generalization across diverse object categories and instance-level variations.

Despite recent progress, current approaches remain inadequate for assembly-oriented spatial reasoning.
Existing assembly methods~\cite{qi2025two,shen2025biassemble,tie2025manual2skill,wu2023leveraging} 
demonstrate promising capabilities within limited object categories and rely on single-modality observations like point clouds, 
which constrains their cross-category generalization at a larger scale.
Spatial reasoning models such as RoboBrain~\cite{ji2025robobrain,team2025robobrain} and RoboRefer~\cite{zhou2026roborefer} incorporate depth-enhanced vision-language representations, yet still lack the geometric precision required for accurate 6D assembly pose prediction. 
The limitations above are further compounded by the scarcity of large-scale multimodal assembly datasets, as existing benchmarks~\cite{ben2021ikea,sener2022assembly101,sliwowski2025reassemble} remain limited in scale and modality coverage, 
hindering the training and evaluation of unified spatial reasoning models.

To address these challenges, we present \textbf{\ours}, a spatial multimodal large language model for robotic assembly.
To overcome the limitations of coarse 2D perception, \ours~employs an $SO(3)$-equivariant point cloud encoder that extracts fine-grained geometric representations.
These 3D features are integrated into a multimodal transformer backbone that jointly processes text, visual manuals, and 3D geometries, enabling explicit spatial reasoning and precise 6D pose prediction.

To address the scarcity of large-scale multimodal data for robotic assembly, we introduce \textbf{AssemBench}, a comprehensive benchmark comprising over 900K multimodal samples derived from 150K distinct assembly steps, spanning diverse object categories including furniture, daily objects, and fragments.
We also develop a data generation pipeline that converts raw mesh assets and web-collected visual resources into high-fidelity point clouds, visual manuals, and aligned linguistic instructions.
This provides large-scale, high-precision supervision for 6D pose reasoning in assembly tasks, enabling models to learn from a diverse range of part geometries and spatial configurations.

Our main contributions are as follows:

\noindent\hangindent=0.8em\hangafter=1
$\bullet$ We propose \textbf{\ours}, a spatial multimodal large language model that bridges raw 3D perception and high-level linguistic reasoning, enabling accurate 6D pose prediction for assembly.

\vspace{-0.4em}
\noindent\hangindent=0.8em\hangafter=1
$\bullet$ We introduce \textbf{AssemBench}, a large-scale multimodal benchmark for assembly-oriented spatial reasoning that includes manuals, point clouds, and textual instructions, comprising over 900K samples with precise 6D pose annotations across diverse assembly categories.

\vspace{-0.4em}
\noindent\hangindent=0.8em\hangafter=1
$\bullet$ We demonstrate state-of-the-art performance across diverse object categories, highlighting strong generalization and real-world robotic assembly potential.

%% file: contents/02_related_work.tex
\vspace{-0.03in}
\section{Related Work}
\vspace{-0.03in}

\textbf{Spatial Reasoning with Large Multimodal Models.}
Perceiving spatial information is fundamental for embodied manipulation.
Prior work mainly falls into two categories: semantic-level reasoning, which captures qualitative spatial relations without metric grounding~\cite{chen2024spatialvlm,cheng2024spatialrgpt}, and numeric-level reasoning, which predicts explicit spatial quantities using RGB-D or multimodal inputs~\cite{cai2025spatialbot,ji2025robobrain,team2025robobrain,zhou2026roborefer}.
However, existing methods struggle with rotation reasoning~\cite{xu2024pointllm} and rely on 2D representations that fail to capture full 3D geometry~\cite{cai2025spatialbot,han2025tiger,ji2025robobrain,liu2026ssr}.
We instead perform equivariant reasoning directly on point clouds for accurate 6D pose estimation.

\textbf{6D Assembly Pose Prediction for Robotic Manipulation.} Accurate 6D pose prediction is essential for assembly execution. 
Existing methods mostly rely on point cloud inputs and $SO(3)$-equivariant representations for geometric reasoning. 
Representative works include $SE(3)$-Transformer~\cite{fuchs2020se}, Vector Neurons~\cite{deng2021vector}, EPN~\cite{chen2021equivariant}, and 3D EGIF~\cite{chen20223d}, which develop equivariant architectures for point clouds and implicit 3D modeling. 
Building on these techniques, $SE(3)$-Assembly~\cite{wu2023leveraging} and TwoByTwo~\cite{qi2025two} show strong performance in assembly pose prediction. 
However, these methods mainly depend on point clouds and generalize within limited assembly domains. 
In contrast, our approach integrates $SO(3)$-equivariant geometric features into a VLM, enabling unified multi-category and multimodal 6D assembly pose prediction with improved generalization.

%% file: contents/03_method.tex
\vspace{-0.03in}
\section{Methods}
\vspace{-0.03in}

\label{sec:method}
\subsection{Problem Formulation}
\label{sec:problem_formulation}
Consider an assembly object $\mathcal{O}=\{o_0,\dots,o_{n-1}\}$ composed of $n$ parts ($n \geq 2$), ordered according to the predefined assembly sequence.
The assembly of the object $\mathcal{O}$ requires 
$n-1$ sequential steps, denoted as $\mathcal{S}=\{s_1,\dots,s_{n-1}\}$.
At step $s_i$, the previously assembled parts are treated as a single fixed entity
$\mathcal{O}^{\text{fixed}}_i = \{o_0, \dots, o_{i-1}\}$, while 
the incoming part
is defined as the moving part $\mathcal{O}^{\text{moving}}_i = \{o_i\}$.
The model's objective is to leverage multimodal inputs to predict a 6D pose representing the rigid transformation required to assemble $\mathcal{O}^{\text{moving}}_i$ onto $\mathcal{O}^{\text{fixed}}_i$.
The model inputs include the point clouds $\mathcal{P}^{\text{fixed}}_i$ and $\mathcal{P}^{\text{moving}}_i$ sampled from $\mathcal{O}^{\text{fixed}}_i$ and $\mathcal{O}^{\text{moving}}_i$, the manual images $I_i^{\text{before}}$ and $I_i^{\text{after}}$ depicting the assembled parts before and after step $s_i$,
and the assembly instruction $l_i$.

\input{contents/fig_model}
\subsection{AssemLM: Multimodal Spatial Reasoning for Robotic Assembly}
\label{sec:model_framework}
We propose AssemLM, a 2B-parameter spatial multimodal large model based on Qwen3-VL-2B-Instruct~\cite{bai2025qwen3}, designed to predict precise 6D poses for sequential assembly steps.
As illustrated in Fig.~\ref{fig:model}, we incorporate four key architectural components: explicit geometric equivariance in 3D perception, embedding-level multimodal fusion for enhanced spatial grounding, a specialized pose tokenizer, and an optimized supervised fine-tuning (SFT) pipeline.

\textbf{Explicit Geometric Equivariance in 3D Perception.}
To empower the LLM backbone with robust spatial reasoning against arbitrary rigid transformations, we inject \textit{$SO(3)$-equivariant representations} into the pipeline. 
Unlike conventional point encoders (e.g., PointNet~\cite{qi2017pointnet}) 
that often discard pose information or rely on data augmentation, 
we employ a specialized Vector Neuron DGCNN~\cite{deng2021vector,wu2023leveraging} to extract features that explicitly track both the orientation and position of assembly parts.

Given a point cloud $P \in \mathbb{R}^{N \times 3}$, we first compute its centered point cloud $P'$ to eliminate the interference of translation on $SO(3)$ equivariance.
We then apply an $SO(3)$-equivariant network to extract rotation-equivariant features $F = \mathcal{E}_{\text{equiv}}(P')$ and fully invariant features $G = \mathcal{E}_{\text{inv}}(P')$. 
For any rotation $R \in SO(3)$, these features mathematically satisfy:
\begin{equation}
\mathcal{E}_{\text{equiv}}(P'R) =  \mathcal{E}_{\text{equiv}}(P') \cdot R,
\end{equation}
\begin{equation}
\mathcal{E}_{\text{inv}}(P'R) = \mathcal{E}_{\text{inv}}(P').
\end{equation}
This formulation ensures that $F$ faithfully preserves the object's 6D pose in the feature space, while $G$ encodes pure geometric shape descriptors independent of spatial placement.
To ground the assembly alignment, we construct a geometric correlation $C$ by modulating the equivariant pose of the moving part ($F_{\text{moving}} \in \mathbb{R}^{f \times 3}$) with the invariant shape context of the fixed part ($G_{\text{fixed}}\in \mathbb{R}^{f \times f}$):
\begin{equation}
C = G_{\text{fixed}} \cdot F_{\text{moving}}, \quad C \in \mathbb{R}^{f \times 3}.
\end{equation}
Since $G_{\text{fixed}}$ is invariant, the correlation $C$ retains the $SO(3)$-equivariance of the moving part. 
This enables the multimodal LLM to capture the spatial state of the moving component relative to the target geometry, facilitating precise 6D pose prediction.
Furthermore,
to resolve non-canonical poses of the fixed part, we incorporate $F_{\text{fixed}}$ as complementary context.
Both $C$ and $F_{\text{fixed}}$ are projected into the LLM embedding space, providing a spatially-grounded geometric prior for assembly planning.

\textbf{Multimodal Fusion and Spatial Grounding.} 
To align manual visuals with 3D geometries, we perform embedding-level fusion to avoid costly cross-attention.
We employ a vision encoder based on SigLIP-2~\cite{bai2025qwen3,dosovitskiy2020image,tschannen2025siglip}
and utilize the \textit{DeepStack mechanism}~\cite{bai2023qwen,bai2025qwen3} to project multi-layer intermediate features into the LLM, effectively preserving high-fidelity spatial semantics.
Concurrently, point cloud embeddings are injected via modality-specific placeholders.
Crucially, applying grid-derived \textit{M-ROPE}~\cite{Qwen2.5-VL} to the preserved visual token layout retains the LLM's inherent spatial reasoning. 
This unified representation supports joint reasoning across text, images, and point clouds, ensuring robust long-context stability for complex assembly tasks.

\textbf{Specialized Assembly Pose Tokenization.} 
We cast assembly pose prediction as a language modeling task by extending the Qwen3-VL vocabulary with 201 discrete tokens to represent coordinate values.
To facilitate smoother model optimization, 6D assembly poses are parameterized as 9D vectors (including 3D translation and a \textit{6D continuous rotation representation}~\cite{zhou2019continuity}), which avoids the singularities of Euler angles and the representation mapping discontinuities caused by the double-cover ambiguity of quaternions.
Each dimension is normalized to $[-1, 1]$ and uniformly discretized into 201 bins. This tokenization strategy offers several advantages: 
\textit{1) Seamless Integration:} It enables an end-to-end autoregressive decoding without requiring auxiliary continuous regression heads or complex pretrained tokenizers. 
\textit{2) Reduced Predictive Burden:} The fixed 9-token sequence length for spatial pose avoids the redundancy and variable-length instability of generic subword tokenizers (e.g., FastTokenizer~\cite{pertsch2025fast}), providing a more stable interface for high-precision grounding. 
\textit{3) Initialization Stability:} The embeddings of new pose tokens are initialized with the mean of existing linguistic embeddings, ensuring training stability during the supervised fine-tuning (SFT) phase.

\textbf{Optimized Supervised Fine-tuning Pipeline.} 
To ensure training stability, we decouple geometric feature acquisition from multimodal reasoning via a two-stage strategy.
In the first \textit{Geometry Warm-up} phase, we pre-train the point cloud encoder alone on an AssemBench subset without linguistic supervision. 
Specifically, an MLP projector is attached to the encoder to directly predict the assembly pose, supervised by an L1 loss for translation and a Geodesic Distance loss for rotation.
This process encourages the encoder to ground its $SO(3)$-equivariant features in pure geometric structures before multimodal fusion.
We then discard the MLP projector and take the pre-trained encoder for the multimodal training stage.
In the subsequent \textit{Full Multimodal Alignment} phase, we jointly optimize all modalities on the complete dataset. 
Notably, we formulate pose prediction as a language modeling task, computing the cross-entropy loss only over the 9-token pose span with all other tokens masked out.
This alignment with the next-token prediction paradigm seamlessly integrates AssemLM into broader embodied AI frameworks.

\input{contents/fig_benchmark}

\subsection{AssemBench}
\label{sec:AssemBench}
\textbf{Overview.}
AssemBench is a comprehensive multimodal benchmark for spatial reasoning in robotic assembly, integrating visual assembly manuals, 3D point clouds, and precise natural language assembly instructions (see Fig.~\ref{fig:dataset_framework}).
Its key features are: 
\textit{1) Large Scale.} The dataset comprises 150K unique assembly steps and over 900K multimodal samples, with diverse visual renderings and instruction granularities to enhance robustness.
Specifically, each step is rendered in three styles: \textit{Freestyle}, \textit{Non-Freestyle}, and \textit{Lineart}. \textit{Freestyle} generates realistic manual-like images but requires a higher rendering cost, while \textit{Lineart} is more efficient with reduced visual detail.
Textual inputs are provided as precise step-level descriptions and vague category-level prompts to enhance data diversity in language instruction.
\textit{2) Rich Diversity.} While existing assembly datasets are limited to categories such as tables and chairs, AssemBench covers over 50 object categories spanning furniture, daily objects, and fragments, fostering broad assembly spatial understanding during SFT. 
\textit{3) High Quality.} High data quality is ensured through deterministic geometric computation for precise 6D pose labels and rejection sampling with multi-model cross-validation for consistent textual instructions.
\textit{4) Easy Scalability.} Our pipeline scales with diverse part-level datasets and web-scale visual assets, enabling continuous expansion.

\textbf{Data Generation.}
Fig.~\ref{fig:dataset_framework} illustrates our progressive data construction pipeline, which combines part-level dataset adaptation with generative expansion to enable general multimodal LLMs to perform assembly-centric spatial reasoning.
\textit{1) Part-Level Mesh Preparation.} 
We curate diverse assembly parts by standardizing existing part-level datasets, including PartNet~\cite{mo2019partnet}, BiAssembly~\cite{shen2025biassemble}, TwoByTwo~\cite{qi2025two}, PartNeXt~\cite{wang2025partnext}, and IKEA-Manual~\cite{wang2022ikea}, into unified formats and coordinate systems. 
We further synthesize part-level meshes with PartPacker~\cite{tang2025efficient} to improve asset diversity.
\textit{2) Assembly Data Generation.}
Given part meshes, we infer valid assembly sequences using spatial coordinates and part connectivity. 
Blender~\cite{blender} is then used to generate step-by-step manuals with relative part poses and corresponding point clouds, providing aligned visual and 3D observations for each assembly step.
\textit{3) Rejection Sampling-Based Semantic Annotation.}
We generate part-level semantic descriptions from manuals and category attributes, and compose step-level instructions using both part semantics and assembly manuals. 
To improve quality, we leverage multiple VLMs to produce candidate annotations, which are judged by GPT-5.2~\cite{singh2025openai} to select the best instruction, ensuring correctness and linguistic diversity. 
More details are provided in Appx.~\ref{appx:AssemBench}.

\input{contents/table_exp_1}

%% file: contents/fig_model.tex
\begin{figure*}[h]
    \centering
    \includegraphics[width=1\linewidth]{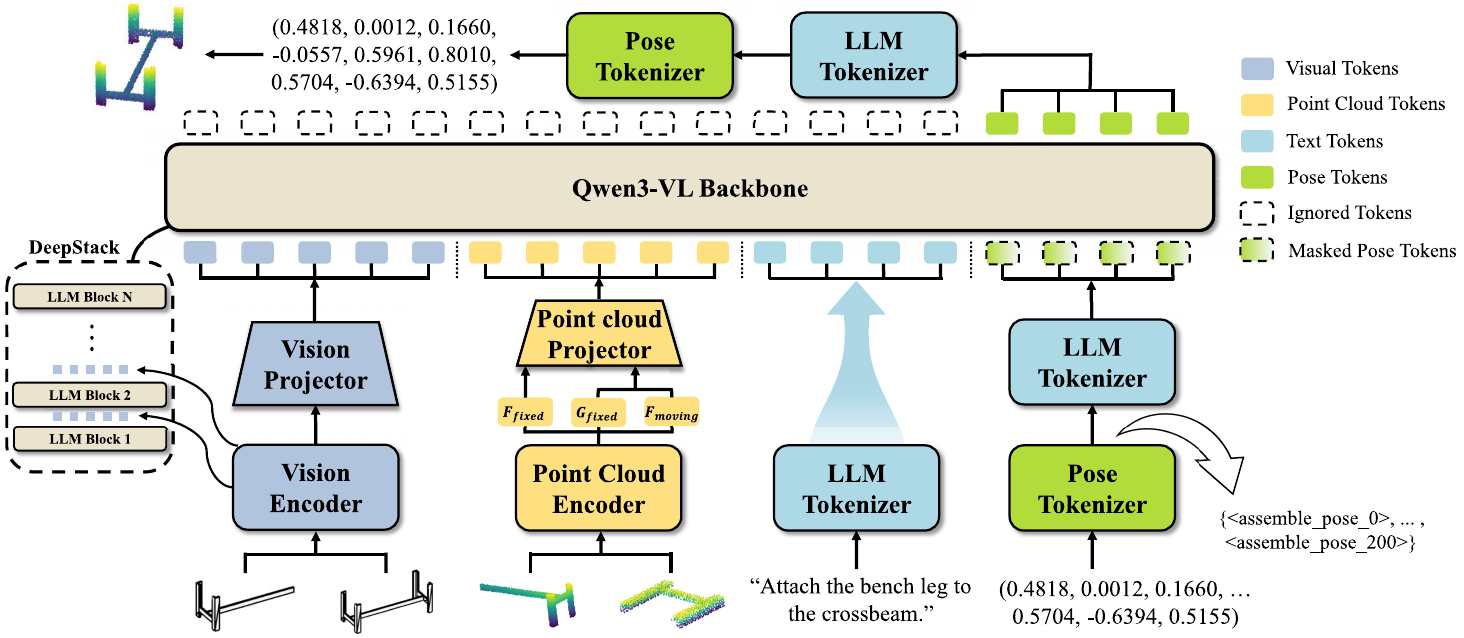}

    \caption{\textbf{Overview of the AssemLM architecture.} 
    AssemLM integrates visual manuals, 3D point clouds, and text instructions into a Qwen3-VL backbone. 
    Visual inputs are projected into the language space, incorporating a DeepStack mechanism for intermediate feature injection. 
    Concurrently, point clouds of fixed and moving parts are processed by an $SO(3)$-equivariant encoder, projecting geometric features ($F$ and $G$) into point cloud tokens. 
    These multimodal embeddings are jointly processed to autoregressively predict discrete pose tokens, 
    subsequently decoded into precise 6D assembly poses.}

    \label{fig:model}
\vspace{-0.18in}
\end{figure*}

%% file: contents/fig_benchmark.tex
\begin{figure*}[t]
    \centering
    \includegraphics[width=1\linewidth]{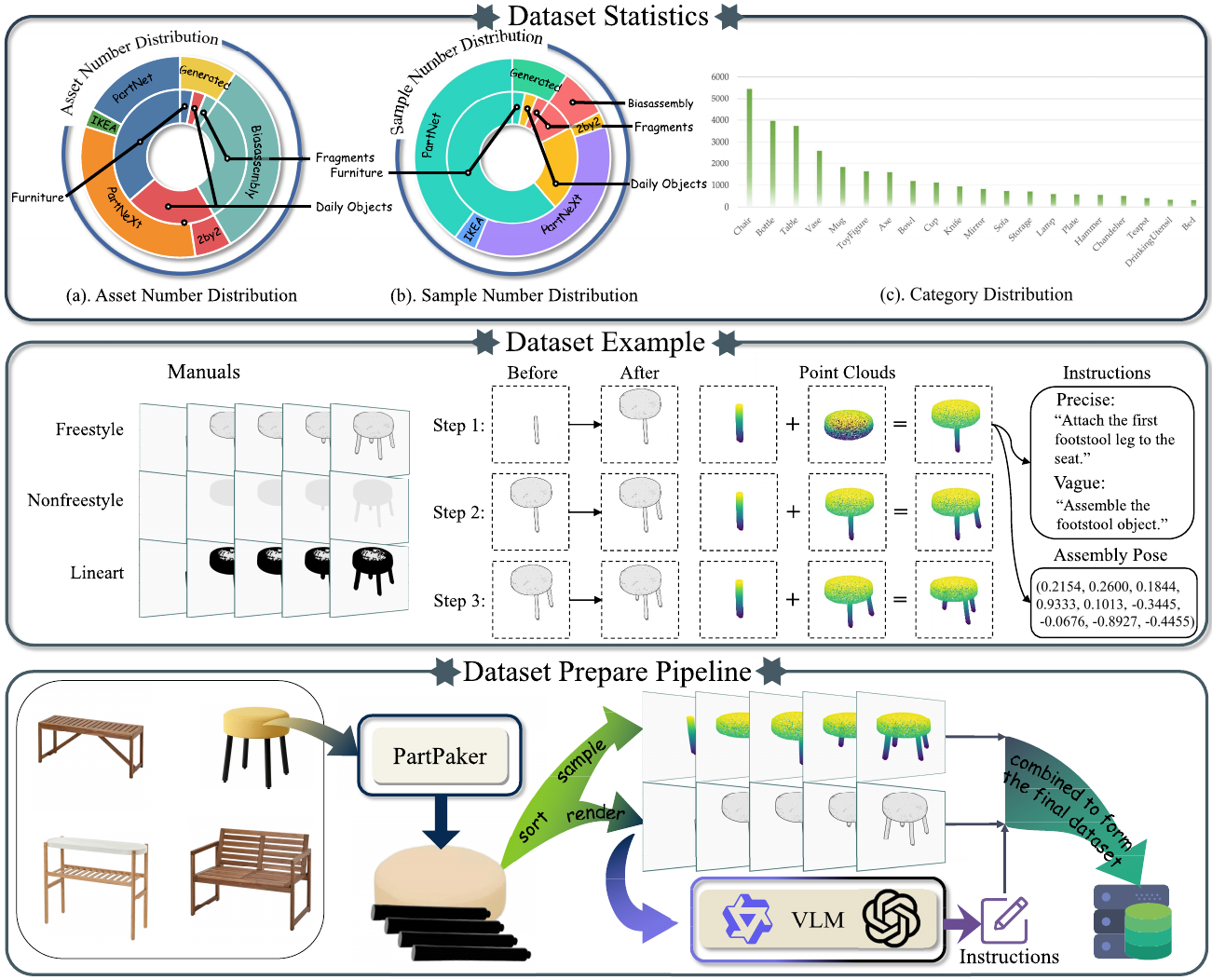}
    \caption{\textbf{Overview of the AssemBench dataset.} 
    (Top) Statistics demonstrating the vast scale and diverse category distributions of AssemBench. 
    (Middle) Multimodal data examples for sequential assembly, including multi-style visual manuals, point clouds, text instructions, and precise 6D poses. 
    (Bottom) Our automated data generation pipeline, which transforms web-sourced image resources into high-fidelity multimodal data to provide large-scale supervision for 6D pose reasoning.}
    \label{fig:dataset_framework}
\vspace{-0.18in}
\end{figure*}

%% file: contents/table_exp_1.tex
\begin{table}[t!] 
\centering
\caption{\textbf{Quantitative Results on Multi-Category Assembly.} AssemLM is compared against baselines (TwoByTwo, GPT-5.2, DeepSeek-V3.2). Metrics: Translation RMSE, Symmetric Chamfer Distance, and Success Rate. The ``All" indicates the weighted average across three macro-categories. Due to space constraints, full individual sub-categories results are provided in Appx.~\ref{appx:Experiments}.
}
\resizebox{1.0\textwidth}{!}{
\begin{tabular}{l|ccc|ccc|ccc|ccc}
\toprule
\multirow{2}{*}{\textbf{Category}} 
& \multicolumn{3}{c|}{\textbf{TwoByTwo~\cite{qi2025two}}} 
& \multicolumn{3}{c|}{\textbf{GPT-5.2~\cite{singh2025openai}}} 
& \multicolumn{3}{c|}{\textbf{DeepSeek-V3.2~\cite{liu2025deepseek}}} 
& \multicolumn{3}{c}{\textbf{AssemLM (Ours)}} \\
& \textbf{RMSE(T)} $\downarrow$ & \textbf{SCD} $\downarrow$ & \textbf{SR} $\uparrow$
& \textbf{RMSE(T)} $\downarrow$ & \textbf{SCD} $\downarrow$ & \textbf{SR} $\uparrow$
& \textbf{RMSE(T)} $\downarrow$ & \textbf{SCD} $\downarrow$ & \textbf{SR} $\uparrow$
& \textbf{RMSE(T)} $\downarrow$ & \textbf{SCD} $\downarrow$ & \textbf{SR} $\uparrow$ \\
\midrule

Bottle
& 0.1953 & 0.1692 & 16.7\%
& 0.2052 & 0.1728 & 19.2\%
& 0.1826 & 0.2564 & 0.0\%
& \textbf{0.0226} & \textbf{0.0102} & \textbf{88.9\%} \\

Plug            
& 0.0483 & 0.0285 & 66.7\%
& 0.2833 & 0.3340 & 0.0\%
& 0.3396 & 0.6262 & 0.0\%
& \textbf{0.0340} & \textbf{0.0053} & \textbf{100.0\%} \\

Postbox         
& 0.2710 & 0.3005 & 0.0\%
& 0.1908 & 0.1823 & 0.0\%
& 0.3050 & 0.5379 & 0.0\%
& \textbf{0.0373} & \textbf{0.0170} & \textbf{78.6\%} \\

Nut             
& 0.0841 & 0.0879 & 12.5\%
& 0.1519 & 0.0830 & 41.7\%
& 0.1964 & 0.2006 & 16.7\%
& \textbf{0.0032} & \textbf{0.0035} & \textbf{100.0\%} \\

Coin            
& 0.1366 & 0.0975 & 50.0\%
& 0.1925 & 0.2541 & 12.5\%
& 0.2500 & 0.4180 & 0.0\%
& \textbf{0.0417} & \textbf{0.0109} & \textbf{87.5\%} \\

Key             
& 0.2624 & 0.3331 & 0.0\%
& 0.1882 & 0.1597 & 0.0\%
& 0.2503 & 0.2695 & 0.0\%
& \textbf{0.0095} & \textbf{0.0038} & \textbf{90.0\%} \\

Usb             
& 0.2496 & 0.2119 & 0.0\%
& 0.3560 & 0.5209 & 0.0\%
& 0.3415 & 0.5744 & 0.0\%
& \textbf{0.0216} & \textbf{0.0050} & \textbf{87.5\%} \\

\midrule
\textbf{Daily Obj.} 
& 0.1504 & 0.1266 & 24.3\%
& 0.1923 & 0.1788 & 17.2\%
& 0.2633 & 0.3350 & 3.6\%
& \textbf{0.0317} & \textbf{0.0096} & \textbf{86.9\%} \\
\midrule

Chair              
& 0.1665 & 0.1391 & 3.6\%
& 0.2718 & 0.3260 & 1.9\%
& 0.2938 & 0.4427 & 1.2\%
& \textbf{0.0161} & \textbf{0.0076} & \textbf{95.6\%} \\

Table              
& 0.2023 & 0.2213 & 1.4\%
& 0.2278 & 0.2904 & 1.3\%
& 0.3262 & 0.6720 & 3.7\%
& \textbf{0.0245} & \textbf{0.0227} & \textbf{82.5\%} \\

Storage
& 0.1400 & 0.1340 & 14.3\%
& 0.1362 & 0.1880 & 5.4\%
& 0.2451 & 0.4409 & 3.6\%
& \textbf{0.0215} & \textbf{0.0205} & \textbf{95.8\%} \\

\midrule
\textbf{Furniture} 
& 0.1761 & 0.1669 & 4.0\%
& 0.2119 & 0.2681 & 2.9\%
& 0.2884 & 0.5185 & 2.8\%
& \textbf{0.0194} & \textbf{0.0139} & \textbf{91.5\%} \\
\midrule

Mirror            
& 0.1982 & 0.1856 & 4.9\%
& 0.2660 & 0.3027 & 4.7\%
& 0.2779 & 0.3818 & 4.3\%
& \textbf{0.0152} & \textbf{0.0179} & \textbf{91.3\%} \\

Plate
& 0.2210 & 0.2089 & 1.8\%
& 0.2747 & 0.3115 & 2.3\%
& 0.2820 & 0.4058 & 2.1\%
& \textbf{0.0120} & \textbf{0.0088} & \textbf{89.7\%} \\

Mug               
& 0.1693 & 0.1414 & 7.3\%
& 0.2664 & 0.3017 & 4.2\%
& 0.2777 & 0.3966 & 3.8\%
& \textbf{0.0115} & \textbf{0.0133} & \textbf{89.0\%} \\

Bowl              
& 0.1922 & 0.1668 & 4.3\%
& 0.2666 & 0.3064 & 4.1\%
& 0.2772 & 0.3980 & 3.7\%
& \textbf{0.0102} & \textbf{0.0099} & \textbf{90.5\%} \\

Wine Glass        
& 0.2036 & 0.1551 & 5.9\%
& 0.2662 & 0.3052 & 4.1\%
& 0.2773 & 0.3972 & 3.7\%
& \textbf{0.0033} & \textbf{0.0074} & \textbf{81.8\%} \\

Teapot            
& 0.1997 & 0.1958 & 7.5\%
& 0.2662 & 0.3060 & 4.2\%
& 0.2775 & 0.3975 & 3.8\%
& \textbf{0.0057} & \textbf{0.0026} & \textbf{95.2\%} \\

Spoon
& 0.0443 & 0.0031 & \textbf{100.0\%}
& 0.3775 & 0.5806 & 0.0\%
& 0.2741 & 0.2774 & 0.0\%
& \textbf{0.0050} & \textbf{0.0003} & \textbf{100.0\%} \\

\midrule
\textbf{Fragments}
& 0.1741 & 0.1420 & 15.1\%
& 0.2670 & 0.3067 & 3.8\%
& 0.2724 & 0.3950 & 3.4\%
& \textbf{0.0097} & \textbf{0.0092} & \textbf{89.8\%} \\
\midrule

\textbf{All}
& 0.1669 & 0.1452 & 14.5\%
& 0.2238 & 0.2512 & 7.9\%
& 0.2747 & 0.4162 & 3.3\%
& \textbf{0.0203} & \textbf{0.0109} & \textbf{89.4\%} \\

\bottomrule
\end{tabular}
}
\vspace{-0.25in}
\label{tab:main_results_multi_category}
\end{table}

%% file: contents/04_experiments.tex
\section{Experiments}
\label{sec:experiment}
In this section, we systematically evaluate AssemLM through three research questions.
\textbf{Q1:} How does \ours compare with state-of-the-art assembly-specific methods and multimodal foundation models? And can it generalize across diverse assembly categories?
\textbf{Q2:} Can AssemLM achieve strong performance on previously unseen dataset sources and novel assembly categories under a zero-shot evaluation setting?
\textbf{Q3:} Is AssemLM's prediction accuracy sufficient for real-world assembly tasks, and how well does it perform on high-precision, multi-step assembly tasks?

We leave additional details in Appx.~\ref{appx:Experiments}, including more baseline comparisons (Appx.~\ref{appx:add_compare}), ablation studies (Appx.~\ref{appx:ablation_study}), 
and experimental details (Appx.~\ref{appx:imp_det} and \ref{appx:real_details}). 

\subsection{Benchmark Comparison on Multi-Category Assembly Tasks}
\label{sec:main_exp}
To evaluate AssemLM’s capacity for unified geometric reasoning, we conduct a comprehensive benchmark comparison across diverse categories, including daily objects, furniture, and fragments.

\textbf{Data Split and Input Standardization.}
We evaluate our method on AssemBench with a 9:1 train-test split. Specifically,
we enforce an instance-level split that keeps all steps of a single object together, and we explicitly isolate the IKEA dataset exclusively for testing.
For fair architectural comparison, we standardize the inputs by utilizing only \textit{Freestyle} manuals and vague instructions.
This avoids redundant multimodal variants, forming a curated training set of ~130K unique state-pose pairs.
Additionally, all input point clouds are centered and $SO(3)$ randomized.

\textbf{Experimental Setup and Baselines.} Unlike category-specific methods, we train a single unified AssemLM on the 130K training set without per-category fine-tuning. 
We compare AssemLM against two baseline categories:
\textit{1) Specialized Assembly Models.} We evaluate TwoByTwo~\cite{qi2025two}, a state-of-the-art framework leveraging $SO(3)$-equivariant representations for high-precision part mating.
\textit{2) Foundation Models.} We include GPT-5.2~\cite{singh2025openai} and DeepSeek-V3.2~\cite{liu2025deepseek} to assess the spatial reasoning capabilities of large-scale models, leveraging their pre-trained spatial commonsense and chain-of-thought reasoning.
Deployment details and further baselines are in Appx.~\ref{appx:imp_det} and~\ref{appx:add_compare}.

\textbf{Evaluation Metrics.}
We evaluate assembly performance using three complementary metrics:
\textit{1) RMSE(T),} which measures translation accuracy via the root mean squared error of the predicted translations.
\textit{2) Symmetric Chamfer Distance (SCD),} explicitly defined as twice the standard Chamfer Distance ($SCD = 2 \times CD$) to capture bidirectional geometric discrepancies and robustly account for object symmetries, bypassing raw rotation errors.
\textit{3) Success Rate (SR),} which denotes the proportion of successful trials, where an assembly is considered successful if $SCD < 0.02$ (i.e., $CD < 0.01$).

\textbf{Results and Analysis.}
As shown in Table~\ref{tab:main_results_multi_category}, AssemLM consistently outperforms baselines across diverse categories, including Furniture, Fragments, and Daily Objects.
Overall, AssemLM achieves an average Success Rate (SR) of 89.4\%, substantially surpassing TwoByTwo (14.5\%).
In terms of precision, AssemLM attains a mean RMSE(T) of 0.0203, which is an order of magnitude lower than that of the competing methods. 
On high-precision tasks such as the \textit{Nut}, \textit{Key}, and \textit{Spoon}, the model reaches near-perfect success rates (90\%--100\%), highlighting its capability to capture the fine-grained geometric constraints required for exact part mating.

Beyond raw performance, the results reveal the limitations of existing approaches. 
Methods relying solely on geometric equivariance (e.g., TwoByTwo) often struggle to scale effectively across large-scale, multi-category datasets.
Meanwhile, foundation models (e.g., GPT-5.2 and DeepSeek-V3.2) yield semantically plausible but spatially imprecise poses, despite structured 3D inputs.
In comparison, AssemLM maintains stable SRs (86.9\%--91.5\%) across evaluated categories, with improved reliability on challenging fragment assemblies. 
These findings suggest that grounding multimodal reasoning in explicit geometric representations facilitates precise and generalizable robotic assembly.

\subsection{Zero-Shot Generalization to Unseen Datasets and Categories}

\label{sec:zero-shot}
To evaluate AssemLM's out-of-distribution (OOD) generalization, we conduct a zero-shot evaluation on the standard IKEA dataset~\cite{wang2022ikea}, which was strictly excluded from training. This assesses whether the geometric reasoning acquired from our randomized 130K dataset can successfully generalize to novel asset sources and categories.

\textbf{Experimental Setup.} Following the configuration in \S\ref{sec:main_exp}, we evaluate on three representative categories: \textit{Bench}, \textit{Chair}, and \textit{Desk}. Notably, \textit{Bench} and \textit{Desk} are entirely unseen during training.

\input{contents/table_exp_2}

\textbf{Results and Discussion.}
As shown in Table~\ref{tab:ikea_results}, AssemLM demonstrates reliable zero-shot generalization. 
Compared to TwoByTwo (6.5\% SR), AssemLM maintains an 81.0\% SR, indicating it learns transferable geometric principles.
It also yields a 0.0263 RMSE(T) under $SO(3)$ randomization, showing steady rotational robustness. 
Conversely, foundation models struggle on IKEA assets ($<5\%$ SR). 
This further validates that without specialized geometric grounding, pre-trained common sense alone cannot resolve the complex spatial configurations required for structural furniture assembly.

\input{contents/fig_overview_real}
\subsection{Real-World Experiments}
\vspace{-0.05in}
\label{sec:rea_exp}

\textbf{Task Design.}
As shown in Fig.~\ref{fig:overview_real}, to evaluate \ours's performance in real-world assembly tasks, we conduct experiments using a Flexiv Rizon 4s robot setup on four challenging tasks:
\emph{Insert Plug}, \emph{Store Cans}, \emph{Insert Flower}, and \emph{Build Blocks}.
Specifically, the \emph{insert plug} and \emph{insert flower} tasks evaluate the model's translation and rotation prediction accuracy 
under stringent spatial tolerances,
while \emph{store cans} and \emph{build blocks} require precise multi-step inference and execution.
See Appx.~\ref{appx:real_details} and Fig.~\ref{fig:real_step_and_manual} for task details and step-wise visualization.

\vspace{-0.01in}
\textbf{Experimental Setup.} 
Building on \S\ref{sec:main_exp}, AssemLM and TwoByTwo are fine-tuned on real-world datasets with randomized spatial perturbations. 
We report execution success rates across 30 random initializations. 
Full construction and implementation details are deferred to Appx.~\ref{appx:detail_real_assets} and \ref{appx:real_details}.

\input{contents/table_real_combine_1}

\textbf{Results and Discussion.}
Table~\ref{tab:real_world_exp} reports the task success rates on four real-world assembly tasks. 
AssemLM consistently outperforms TwoByTwo across all tasks, indicating 
superior robustness and generalization to physical assets.
In particular, 
AssemLM doubles the success rate on \emph{Insert Plug}, highlighting improved fine-grained pose reasoning under tight precision constraints, and excels in multi-step tasks (\emph{Store Cans} and \emph{Build Blocks}).
For \emph{Insert Flower}, both models achieve high success given the lower translational precision required; failures primarily result from initial perturbations causing long-stem collisions with the vase.

\textbf{Step-wise Analysis.} Table~\ref{tab:real_world_exp2} details multi-step execution for \emph{Store Cans}. 
While overall success declines due to sequential error accumulation, AssemLM achieves an average per-step success rate of 
89.3\%.
We attribute this robustness to its effective multimodal fusion: even under partial occlusion (Fig.~\ref{fig:real_step_and_manual}), AssemLM extracts critical cues from manuals to compensate for the sparse point clouds. 
Conversely, TwoByTwo repeatedly targets previously occupied slots, failing to track evolving task states. 
Ultimately, AssemLM's failures stem primarily from long-horizon compounding errors rather than isolated pose inaccuracies, underscoring its reliable step-level spatial reasoning.

%% file: contents/table_exp_2.tex
\begin{table*}[ht]
\vspace{-0.10in}
\centering
\caption{\textbf{Zero-Shot Generalization on the IKEA Dataset.}
Quantitative results for out-of-distribution (OOD) evaluation.
\textit{Bench} and \textit{Desk} represent unseen categories.
Metrics: Translation RMSE, Symmetric Chamfer Distance, and Success Rate. The ``All" indicates unweighted average.}
\resizebox{1.0\textwidth}{!}{%
\begin{tabular}{l|ccc|ccc|ccc|ccc}
\toprule
\multirow{2}{*}{\textbf{Category}}
& \multicolumn{3}{c|}{\textbf{TwoByTwo}}
& \multicolumn{3}{c|}{\textbf{GPT-5.2}}
& \multicolumn{3}{c|}{\textbf{DeepSeek-V3.2}}
& \multicolumn{3}{c}{\textbf{AssemLM (Ours)}} \\
& \textbf{RMSE(T)} \(\downarrow\) & \textbf{SCD} \(\downarrow\) & \textbf{SR} \(\uparrow\)
& \textbf{RMSE(T)} \(\downarrow\) & \textbf{SCD} \(\downarrow\) & \textbf{SR} \(\uparrow\)
& \textbf{RMSE(T)} \(\downarrow\) & \textbf{SCD} \(\downarrow\) & \textbf{SR} \(\uparrow\)
& \textbf{RMSE(T)} \(\downarrow\) & \textbf{SCD} \(\downarrow\) & \textbf{SR} \(\uparrow\) \\
\midrule

Bench
& 0.1826 & 0.1983 & 9.9\%
& 0.2213 & 0.3435 & 1.4\%
& 0.3304 & 0.5922 & 0.0\%
& \textbf{0.0330} & \textbf{0.0250} & \textbf{81.4\%} \\

Chair
& 0.1810 & 0.1695 & 7.1\%
& 0.2189 & 0.2661 & 5.4\%
& 0.2627 & 0.4273 & 1.7\%
& \textbf{0.0205} & \textbf{0.0164} & \textbf{79.5\%} \\

Desk
& 0.2401 & 0.2973 & 2.5\%
& 0.1957 & 0.2585 & 7.5\%
& 0.2989 & 0.5274 & 5.1\%
& \textbf{0.0255} & \textbf{0.0225} & \textbf{82.1\%} \\

\midrule
\textbf{All}
& 0.2012 & 0.2217 & 6.5\%
& 0.2120 & 0.2893 & 4.8\%
& 0.2973 & 0.5156 & 2.3\%
& \textbf{0.0263} & \textbf{0.0213} & \textbf{81.0\%} \\

\bottomrule
\end{tabular}%
}
\label{tab:ikea_results}
\vspace{-0.05in}

\end{table*}

%% file: contents/fig_overview_real.tex
\begin{wrapfigure}{r}{0.5\textwidth}
    \centering
    \vspace{-13pt}
    \includegraphics[width=1\linewidth]{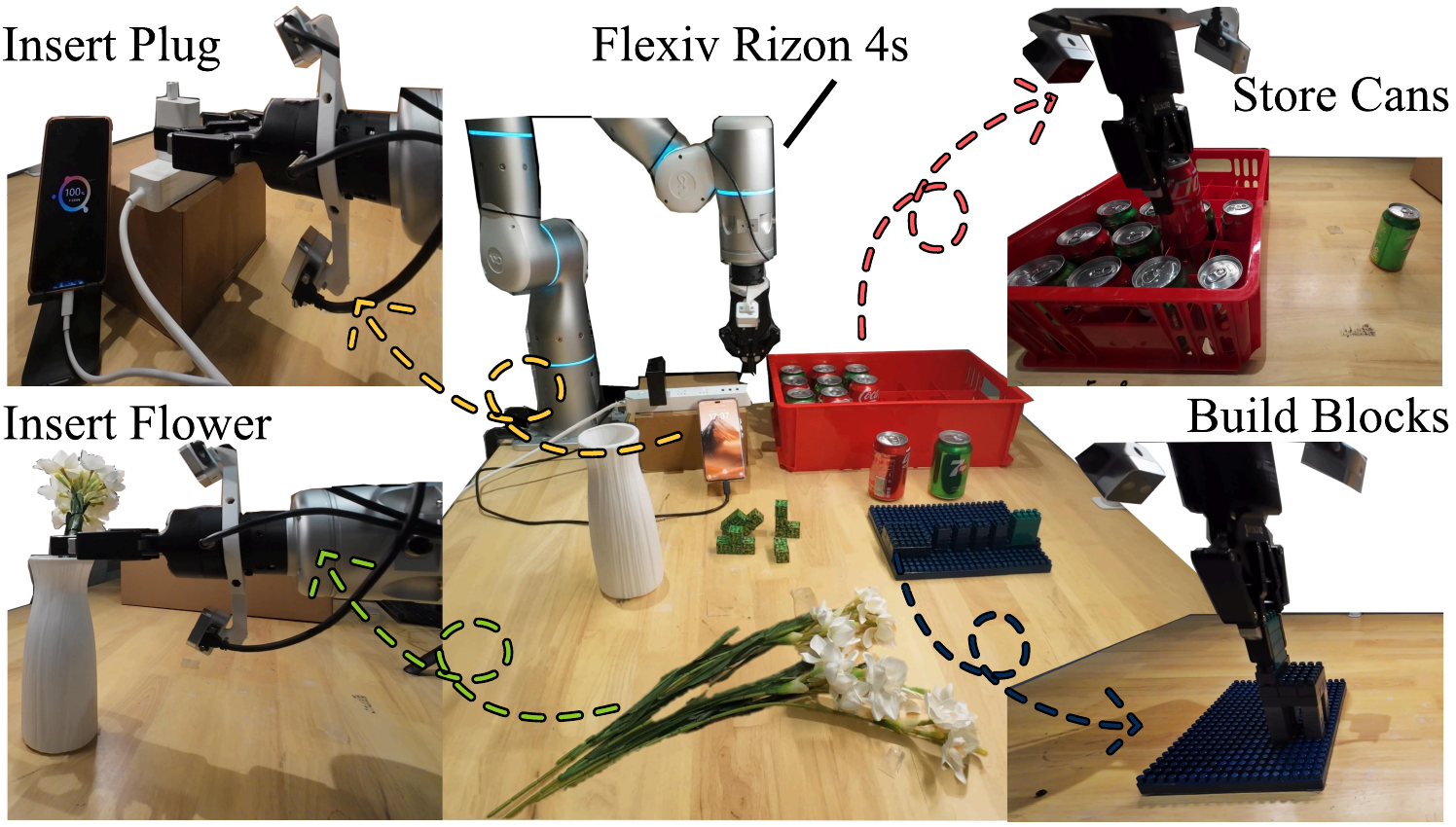}
    \vspace{-13pt}
    \caption{\textbf{Real-world setup.} We evaluate four challenging tasks using a Flexiv Rizon 4s arm.}
    \vspace{-12pt}
    \label{fig:overview_real}
\end{wrapfigure}

%% file: contents/table_real_combine_1.tex
\begin{table}[htb]
    \vspace{-0.08in}
    \centering
    \begin{minipage}[t]{0.48\textwidth}
        \centering
        \caption{Real-world assembly success rates.}
        \resizebox{\linewidth}{!}{
            \begin{tabular}{lccccc}
            \toprule
            \textbf{Task} &\textbf{Plug}& \textbf{Cans} & {\textbf{Flower}} & \textbf{Blocks}  \\
            \midrule
            \textbf{TwoByTwo~\cite{qi2025two}} & $6/30$ &$0/30$ & $19/30$ & $4/30$   \\
            \textbf{AssemLM (Ours)} &$13/30$ & $19/30$ & $20/30$ & $9/30$ \\
            \bottomrule
            \end{tabular}
        }
        \label{tab:real_world_exp}
    \end{minipage}
    \hfill
    \begin{minipage}[t]{0.48\textwidth}
        \centering
        \caption{Cumulative successes on \emph{Store Cans}.}
        \resizebox{\linewidth}{!}{
            \begin{tabular}{lcccc}
            \toprule
            \textbf{Step Number} &\textbf{Step 1}& \textbf{Step 2} & {\textbf{Step 3}} & \textbf{Step 4}  \\
            \midrule
            \textbf{TwoByTwo~\cite{qi2025two}} & $17/30$ &$2/30$ & $0/30$ & $0/30$  \\
            \textbf{AssemLM (Ours)} &$25/30$ & $23/30$ & $20/30$ & $19/30$ \\
            \bottomrule
            \end{tabular}
        }
        \label{tab:real_world_exp2}
    \end{minipage}
    \vspace{-0.08in}
\end{table}

%% file: contents/05_conclusion.tex
\vspace{-0.05in}
\section{Limitations}
\vspace{-0.05in}
Our work has the following limitations. 
First, while \ours~demonstrates robust spatial reasoning, its performance can degrade when processing severely noisy or highly occluded point clouds directly sampled from real-world depth cameras.
Second, while the model shows promising generalization in both simulation and real-world settings, it requires additional perception post-processing in real deployments to achieve optimal performance.
Finally, the current model is mainly designed for rigid assembly tasks, and future work could explore more flexible architectures that generalize to a wider range of embodied reasoning and manipulation scenarios.
\vspace{-0.05in}
\section{Conclusion}
\vspace{-0.05in}
In this work, we present AssemLM, pioneering the use of Multimodal LLMs for precise pose reasoning in robotic assembly, pushing the boundaries of spatial reasoning in embodied foundation models. To bridge semantic intent and geometric execution, our architecture intrinsically couples $SO(3)$-equivariant point clouds with a vision-language backbone, stabilized by a discrete pose tokenizer. We construct AssemBench via an automated pipeline, yielding over 900K multimodal samples and 150K assembly steps. Evaluations confirm state-of-the-art performance, robust zero-shot generalization in simulation, and 
task-adapted real-world assembly execution.
Ultimately, AssemLM establishes a powerful paradigm for geometry-aware, language-driven robotic manipulation.

\section*{Acknowledgments}
This work is supported by the National Natural Science Foundation of China (Grant Nos. 62427819 and 62306242),  the Young Elite Scientists Sponsorship Program by CAST (Grant No. 2024QNRC001), and the Yangfan Project of the Shanghai (Grant No. 23YF11462200), and the Science and Technology Commission of Shanghai Municipality (No. 24511103100).

%% file: contents/99_appendix.tex
\clearpage
\newpage

\begin{appendix}
\section{Implementation Details of AssemLM}
\label{appx:AssemLM}

\subsection{Generalization to Real-World Assets}
\label{appx:detail_real_assets}
\input{contents/fig_real_assets}
To evaluate the transferability of AssemLM, we extend our framework to physical objects by leveraging the unique characteristics of real-world data acquisition. Unlike web-scale images that are often restricted to a single viewpoint and monolithic object representations, real-world assets provide two distinct advantages for geometric reasoning. 
First, multi-view image acquisition ensures the completeness of the generated 3D object assets, effectively mitigating geometric inconsistencies caused by self-occlusions.
Second, this approach allows for independent, fine-grained 3D asset generation for each individual part, ensuring high-fidelity local geometry. This comprehensive and detailed representation successfully narrows the sim-to-real gap during physical deployment.

As illustrated in Fig.~\ref{fig:real_assts}, we employ the Hunyuan3D~\cite{zhao2025hunyuan3d} model to generate high-fidelity 3D models for each part of the task-specific assets used in our experiments. Following the automated dataset construction pipeline described in Appx.~\ref{appx:AssemBench}, these real-world assets are further organized into a structured dataset format suitable for inference with AssemLM.

In a full deployment pipeline, while front-end perception modules for 6D pose estimation~\cite{wen2024foundationpose} and grasp prediction~\cite{fang2020graspnet, fang2023anygrasp} have matured to reliably handle 
object localization and grasping,
the primary bottleneck remains determining the exact target pose of the grasped object required for physical assembly.
To bridge this gap, AssemLM successfully addresses this critical reasoning challenge.
Consequently, our framework structurally complements these established perception front-ends and standard low-level motion planners, serving as the core reasoning engine to complete a real-world robotic assembly pipeline spanning perception, reasoning, and execution.

More importantly, the key assembly poses predicted by AssemLM offer broader utility beyond standalone deployment. 
Providing these precise spatial configurations effectively resolves a shared bottleneck in both advanced manipulation frameworks~\cite{huang2024rekep,pan2025omnimanip} and simulation data generation pipelines~\cite{chen2025robotwin,jing2025humanoidgen}, where determining the exact target pose for object placement remains a central challenge.

\subsection{Special Tokens and Chat Template Design}
As described in \S\ref{sec:model_framework}, to equip the language model backbone with the ability to process multimodal inputs, we extend Qwen3-VL by introducing special tokens specifically designed for point cloud information. 
Specifically, each element of the 9D pose vector is normalized to the range $[-1, 1]$ and uniformly discretized into $201$ bins, which are mapped to a set of continuous tokens $\{\texttt{<assemble\_pose\_0>}, \dots, \texttt{<assemble\_pose\_200>}\}$. Here, $\texttt{<assemble\_pose\_0>}$ and $\texttt{<assemble\_pose\_200>}$ correspond to $-1$ and $1$, respectively. This tokenization strategy yields a precise bin width of $0.01$, bounding the maximum quantization error within $0.005$, which rigorously satisfies the high-precision requirements of robotic assembly. Furthermore, we introduce $\texttt{<pointcloud>}$ as a structural placeholder for point cloud features, enclosed by $\texttt{<PC\_START>}$ and $\texttt{<PC\_END>}$ to explicitly demarcate the boundaries of the geometric sequence.

\subsection{Model Architecture Details}
Following Qwen3-VL-2B, our \textbf{vision encoder} adopts the SigLIP-2 architecture, specifically SigLIP2-Large (300M). 
It comprises a patch embedding layer, positional encoding modules, and 24 stacked vision transformer blocks, with a hidden size of 1024, an MLP intermediate dimension of 4096, and 16 attention heads. 
We further extract intermediate visual features from vision blocks 5, 11, and 17, process them with three DeepStack merger modules (each consisting of a LayerNorm and a two-layer MLP), and inject them into the first three transformer layers of the large language model backbone.

For the \textbf{point cloud encoder}, Vector Neuron DGCNN produces an $SO(3)$-equivariant feature $\mathbf{F} \in \mathbb{R}^{1024 \times 3}$ and an $SO(3)$-invariant feature $\mathbf{G} \in \mathbb{R}^{1024 \times 1024}$. 
Accordingly, the feature hyperparameter $f$ introduced in \S\ref{sec:model_framework} is set to 1024. After the \textbf{modality projectors}, the two manual images and the two part point clouds are embedded into representations of size $(648, 2048)$ and $(512, 2048)$, respectively, before being injected into the transformer backbone.

\section{AssemBench Dataset and Production Pipeline}
\label{appx:AssemBench}
\input{contents/fig_canonical_coordinate_system}
As mentioned in \S\ref{sec:AssemBench}, we generate a precise and logically ordered assembly sequence for each asset to ensure geometric validity
and correct part connectivity throughout the assembly process. 
To satisfy the training requirements described in \S\ref{sec:problem_formulation}, we further construct multimodal supervision for each assembly step in the sequence, including high-fidelity point clouds, structured assembly manuals, and corresponding textual instructions (see Fig.~\ref{fig:daily_frag}). 
In the following, we describe the dataset in detail, organized according to the data generation procedure.

\subsection{Asset Normalization and Canonical Coordinate System}
\input{contents/fig_real_step_and_manual}
To standardize all assets for downstream processing and model inference, we define a \textit{canonical coordinate system} that unifies object scale and spatial placement, as illustrated in Fig.~\ref{fig:canonical_coordinate_system}. 
Specifically, we first rotate each asset into a consistent \textit{z}-up orientation. 
For example, assets from the IKEA-Manual~\cite{wang2022ikea} are originally defined in a \textit{y}-up coordinate system and are therefore rotated accordingly. 
We then normalize the object scale by rescaling the longest axis to unit length, translate the object such that its center is aligned with the origin in the horizontal plane, and place the lowest point of the object at zero height along the vertical axis. 
This normalization procedure yields a physically plausible configuration that mimics objects resting on the ground in real-world scenes, while providing a unified geometric representation for learning and reasoning.

\subsection{Point Cloud Sampling and Assembly Order Verification}
After transforming all assets into the canonical coordinate system, we perform surface point cloud sampling for each part.
Specifically, we first apply area-weighted surface sampling using the \textit{trimesh} library to randomly sample 10,240 surface points per part, and then employ \textit{Farthest Point Sampling (FPS)} to downsample them to 1,024 points.
Compared to using area-weighted sampling alone, this two-stage strategy yields a more uniformly distributed and geometrically representative point cloud, preserving critical local structures such as sharp edges and fine connectors that are essential for accurate assembly reasoning and downstream 6D pose estimation.

Using the sampled point clouds of each part, we construct a binary connectivity matrix $M$, where $M_{ij}$ indicates whether parts $o_i$ and $o_j$ are physically connected. 
The part $o$ definition follows \S\ref{sec:problem_formulation}. Specifically, two parts $o_i$ and $o_j$ are considered connected if the minimum Euclidean Distance between their corresponding surface point clouds $P_i$ and $P_j$ is below a predefined threshold $\tau$, i.e.,
\begin{equation}
M_{ij} = \mathbb{I}\!\left( \min_{p\in P_i,\;q\in P_j} \lVert p-q \rVert_2 < \tau \right), \quad \tau = 0.06.
\end{equation}

We maintain two dynamic part sets: an assembled set $\mathcal{O}^{a}$ and an unassembled set $\mathcal{O}^{u}$. 
Given an assembly consisting of $n$ parts with an ordering $\{o_0, \dots, o_{n-1}\}$, each assembly step $s_i$ is abstracted as selecting a part $o_i$ from $\mathcal{O}^{u}_i$ and moving it into $\mathcal{O}^{a}_i$, yielding updated sets $\mathcal{O}^{a}_{i+1}$ and $\mathcal{O}^{u}_{i+1}$. 
This process iterates until all parts are assembled.

At the initial step ($i=0$), we select the base part with the lowest vertical extent:
\begin{equation}
o_0 = \arg\min_{o \in \mathcal{O}^{u}_0} \;\min_{p \in P_o} \; p_z,
\end{equation}
where $p_z$ denotes the $z$-coordinate of point $p$. 

For subsequent steps ($i \geq 1$), we restrict candidate parts to those connected to the current assembled set according to $M$, and select the one with the lowest vertical extent:
\begin{equation}
o_i = \arg\min_{o \in C_i} \;\min_{p \in P_o} \; p_z, 
C_i = \left\{ o \in \mathcal{O}^{u}_i \;\middle|\; \exists\, o' \in \mathcal{O}^{a}_i,\; M_{oo'} = 1 \right\}.
\end{equation}

This strategy enforces part connectivity while assembling components in a bottom-up manner, thereby ensuring both structural validity and geometric feasibility throughout the assembly process.

Based on the determined assembly order and the sampled point clouds, we derive the fixed and moving point cloud representations for each assembly step, denoted as $\mathcal{P}^{\text{fixed}}_i$ and $\mathcal{P}^{\text{moving}}_i$, respectively, as defined in \S\ref{sec:problem_formulation}.
Specifically, the moving point cloud is given by
\begin{equation}
\mathcal{P}^{\text{moving}}_i = P_i,
\end{equation}
while the fixed point cloud aggregates all previously assembled parts.
To maintain a consistent point cloud resolution across steps, we further apply FPS to downsample the union of fixed parts to 1024 points:
\begin{equation}
\mathcal{P}^{\text{fixed}}_i = \mathrm{FPS}\!\left( P_0 \cup \cdots \cup P_{i-1} \right).
\end{equation}

\subsection{Instruction Manual Generation}
For each assembly step $s_i$, we provide a pair of rendered images before and after the assembly, denoted as $I_i^{\text{before}}$ and $I_i^{\text{after}}$, which serve as visual instruction manuals to guide the model in understanding the intended assembly operation.
These images provide step-specific spatial cues that complement the geometric information from point clouds, enabling the model to infer the correct 6D assembly pose.
Such visual guidance is particularly important for tasks with multiple valid attachment locations, e.g., selecting the correct mounting position for one of several identical furniture legs, or determining the appropriate placement slot in real-world storage tasks such as arranging cans in a basket.

To generate the instruction manuals, we sequentially insert the normalized assets $\{o_0, \dots, o_{n-1}\}$ into a Blender~\cite{blender}
scene following the determined assembly order.
This process produces a sequence of rendered images $\{I_0, \dots, I_{n-1}\}$, where for assembly step $s_i$, the pre- and post-assembly images are defined as
$I_i^{\text{before}} = I_i$ and $I_i^{\text{after}} = I_{i+1}$, satisfying
$I_i^{\text{after}} = I_{i+1}^{\text{before}}$.
This formulation ensures temporal consistency between consecutive assembly steps.

To enrich the visual diversity of the dataset, we render each assembly step using three complementary rendering styles: \textit{Freestyle}, \textit{Non-Freestyle}, and \textit{Lineart} (see Fig.~\ref{fig:furniture1}).
The \textit{Freestyle} style performs a topology-aware edge detection after photorealistic rendering, producing line drawings with fine structural details that highlight critical assembly features.
While highly informative, this mode incurs a higher rendering cost, with an average rendering time of 8.59 seconds per image, depending on scene complexity and Cycles sampling settings.
The \textit{Non-Freestyle} style disables edge extraction and yields standard pixel-based renderings, offering faster rendering at an average of 3.35 seconds per image, albeit with less explicit structural emphasis.
The \textit{Lineart} style leverages Blender's Grease Pencil system and its Line Art modifier to algorithmically extract feature lines directly in 3D space and project them into vectorized 2D drawings.
This approach achieves significantly faster rendering (0.22 seconds per image on average), though the resulting visuals are coarser and may contain noise.
Together, these complementary rendering styles balance visual fidelity, structural clarity, and computational efficiency, providing diverse and informative supervision for multimodal assembly reasoning.

\section{Experimental Details and Supplementary Experiments}
\label{appx:Experiments}
\input{contents/table_ablition}
\input{contents/table_addition_compare}
\subsection{Implementation Details}
\label{appx:imp_det}
For the benchmark comparison experiments in \S\ref{sec:main_exp}, the zero-shot generalization experiments in \S\ref{sec:zero-shot}, as well as the supplementary baseline comparisons in Appx.~\ref{appx:add_compare} and ablation studies in Appx.~\ref{appx:ablation_study}, all trainable models are trained on the same 130K training samples for 12 epochs under identical settings. 
Specifically, we adopt the same hyperparameters (e.g., \textit{per\_gpu\_batch\_size}=4 and temperature=0) and the same training environment consisting of 4$\times$A100/H100 GPUs, each with 80GB memory. For the foundation models, we serialize sampled point coordinates into structured text tokens to facilitate spatial reasoning. Concretely, for each part, we sample 1024 points and represent them as comma-separated 3D coordinate tuples in the form of $(x,y,z)$. These coordinates are combined with the before/after assembly manuals and the task instruction to form the multimodal prompt, and the model is required to predict the 9D pose vector in list form, parameterized by 3D translation and a 6D continuous rotation representation. All input point clouds are normalized and augmented with random $SO(3)$ rotations. Moreover, the same set of randomized initial poses is used across all evaluations to ensure a fair and consistent comparison.

\input{contents/list_prompt}

As detailed in \S\ref{sec:main_exp}, we partition our evaluation suite using a rigorous object-level splitting strategy to prevent data leakage. Specifically, we guarantee that all assembly steps associated with a single object reside exclusively within either the training or the testing set. 
For datasets with predefined train-test splits, such as the TwoByTwo dataset~\cite{qi2025two}, we strictly adhere to their official training and testing partitions.
For datasets lacking native partitions, including our automatically generated dataset, we adopt a 9:1 train-to-test ratio. Crucially, to evaluate the cross-dataset generalization capability of AssemLM, assets from IKEA-Manual~\cite{wang2022ikea} are reserved exclusively for testing.

\subsection{Additional Experimental Setup}
As shown in Tables~\ref{tab:ablition} and~\ref{tab:additional_baseline}, DO, Fur., Frag., and R-T denote Daily Objects, Furniture, Fragments, and RMSE(T), respectively, following the same conventions as in \S\ref{sec:main_exp} and \S\ref{sec:zero-shot}. To further validate the advantages of AssemLM, we introduce two additional experimental settings. First, compared with \S\ref{sec:zero-shot}, we extend Fur.* to the full IKEA-Manual dataset~\cite{wang2022ikea} by additionally including the \emph{shelf}, \emph{table}, and \emph{misc} categories, where \emph{misc} consists of a mixture of uncategorized furniture items. This setting poses a more challenging test of the model's reasoning ability on unseen categories and assets from unseen sources. Second, to better assess rotational prediction, we introduce the metric SCD(R), which measures the Chamfer Distance induced solely by rotational error under accurate translation. We do not adopt Geodesic Distance or RMSE(R) because many assembly parts are symmetric, such that multiple rotations can be equally valid, and these metrics may therefore over-penalize correct predictions.

\input{contents/table_exp_1_full}

\subsection{Ablation Studies}
\label{appx:ablation_study}
To validate the contributions of different input modalities and architectural components, we conduct both modality and module ablations.

For modality ablation, we train and evaluate the model by independently removing either the visual manuals or the textual instructions, and report the corresponding results in the top part of Table~\ref{tab:ablition}.
Removing either the visual or textual information leads to elevated estimation errors in both translation and rotation. Notably, the exclusion of visual features causes a particularly pronounced increase in translation error, demonstrating that manual illustrations provide key spatial guidance for anchoring precise target locations, which ultimately underscores that multimodal synergy is vital for high-precision pose prediction.

For module ablation, the results in the bottom part of Table~\ref{tab:ablition} show that removing each key design component leads to performance degradation to different extents. This confirms that each module contributes positively to the overall performance of AssemLM.

\subsection{Additional Comparative Experiments}
\label{appx:add_compare}
We additionally include ManualPA~\cite{zhang2025manual}, which predicts assembly actions from manuals and point clouds, and SE(3)-Assembly~\cite{wu2023leveraging}, which predicts assembly poses using SO(3)-equivariant representations, as supplementary baselines. As shown in Table~\ref{tab:additional_baseline}, our method achieves higher prediction accuracy than both baselines, with particularly notable gains in translation estimation, where it substantially outperforms ManualPA.

\subsection{Additional Details of Real-World Experiments}
\label{appx:real_details}
For the real-world experiments in \S\ref{sec:rea_exp}, we design four challenging tasks that cover both fine-grained manipulation and multi-step assembly to validate the effectiveness of our model. The four tasks, including \emph{Insert Plug}, \emph{Store Cans}, \emph{Insert Flower}, and \emph{Build Blocks}, are described in detail below:

\begin{itemize}[leftmargin=*, labelsep=0.5em]
\item \textit{Insert Plug.}  
The robot is required to insert a charger plug into a standard power strip to initiate phone charging. Unlike prior works that introduce adapters or enlarged sockets to increase tolerance~\cite{qi2025two}, we directly use a real charger plug with metal contacts of size $0.6\,\text{cm} \times 0.1\,\text{cm,}$ and a standard socket opening of $0.8\,\text{cm} \times 0.2\,\text{cm}$. Successful execution requires controlling the lateral alignment error within $0.1\,\text{cm}$ and fully inserting the $1.5\,\text{cm}$-deep plug. This task places extremely strict demands on the model’s 6D pose prediction accuracy in both translation and rotation.

\item \textit{Store Cans.}  
This task evaluates multi-step reasoning and execution accuracy. The robot sequentially places four identical cans into distinct compartments of a plastic basket. A single failure in any step causes the entire task to fail. Each can has a circular base of $7\,\text{cm}$ diameter, while each target compartment is a $7.5\,\text{cm} \times 7.5\,\text{cm}$ square, leaving minimal clearance. Moreover, due to the large spatial extent of the 
basket, point cloud sampling under a fixed budget (1024 points) yields sparse observations of the cans within the $\mathcal{P}^{\text{fixed}}$ representation. 
Since all $\mathcal{P}^{\text{moving}}$ are geometrically similar, the model must rely on subtle geometric differences in sparse point clouds to infer slot occupancy and reason about the valid target slot for the current step.

\item \textit{Insert Flower.}  
In this task, the robot inserts a flower into a vase while maintaining the vase's stability (see Fig.~\ref{fig:real_assts}). We select flowers with significant stem curvature: although bent, the flower height reaches $45\,\text{cm}$, nearly twice the vase height of $23.2\,\text{cm}$. The horizontal projection of the curved stem spans approximately $20\,\text{cm}$, while the vase opening diameter is only $6\,\text{cm}$. Accurate rotation prediction is therefore critical; even small orientation errors can cause lateral displacement of the stem, resulting in collision and tipping of the vase.

\item \textit{Build Blocks.}  
This task requires assembling a block-based chair in two sequential steps. First, the robot inserts the seat component onto four vertical legs simultaneously. Each peg has a diameter of only $0.5\,\text{cm}$, and all four contact points must align precisely; otherwise, early contact induces lateral forces that displace the remaining legs, leading to failure. The second step inserts the backrest onto the assembled seat, where any residual error from the first step accumulates and further amplifies task difficulty. This task strongly challenges both translational and rotational precision as well as error accumulation across steps.
\end{itemize}

During fine-tuning and evaluation of AssemLM and TwoByTwo on datasets constructed from real-world assets, we apply translational perturbations within a range of 0.1 on the horizontal plane in the canonical coordinate system, together with rotational perturbations of up to 10$^\circ$, to simulate randomized initial conditions. All models are fine-tuned for 330 epochs with a batch size of 4 and evaluated under 30 randomized initializations with seeds 0--29.
In experiments, the grasp poses are predefined, while the assembly poses are predicted by the model, thereby mitigating compounding errors caused by imprecise object localization or grasping.

\input{contents/fig_ablition}

\subsection{Further Analysis on Data and Design Choices}
To further investigate the factors affecting AssemLM's performance, we conduct a set of additional analyses on the TwoByTwo testing partition~\cite{qi2025two}, specifically evaluating the impact of dataset scale, rotation randomization range, and tokenizer design on geometric reasoning. We consider four experimental settings to examine these factors in a controlled manner, as detailed below:

\begin{itemize}[leftmargin=*, labelsep=0.5em]
\item \textit{AssemLM (Full).}
The proposed model trained on the full curated dataset of 130K samples with $SO(3)$ rotation randomization.

\item \textit{AssemLM (350-S).}
A low-data variant trained exclusively on 350 samples from the daily object subset of the TwoByTwo training split, designed to evaluate data efficiency under constrained data regimes.

\item \textit{AssemLM (Lim-Rot).}
A constrained variant where the rotation randomization is restricted to $\pm 45^{\circ}$ during both training and testing, assessing the model's sensitivity to pose distribution.

\item \textit{AssemLM (Fast-Tk).}
A variant utilizing the FastTokenizer architecture instead of our custom pose tokenizer, aimed at evaluating the effectiveness of our discretization strategy.
\end{itemize}

As illustrated in Fig.~\ref{fig:ablation_analysis}, quantitative results indicate that the scale of the training data is the most decisive factor for generalization. \textit{AssemLM (350-S)}, limited to only 350 samples, fails to capture the underlying geometric manifolds, resulting in a high average RMSE(T) of 0.0475. In contrast, \textit{AssemLM (Full)} leverages the 130K samples to achieve a significantly lower error of 0.0317, demonstrating that large-scale multimodal pre-training is essential for learning universal assembly priors. Regarding rotation randomization, while \textit{AssemLM (Lim-Rot)} achieves localized precision in constrained settings, it offers no significant advantage over the \textit{Full} model, suggesting that full rotation training does not compromise local accuracy while enabling broader generalization.

Furthermore, the comparison between \textit{AssemLM (Full)} and \textit{AssemLM (Fast-Tk)} highlights the importance of the tokenizer design. We observe a distinct performance degradation in the FastTokenizer variant, with RMSE(T) increasing to 0.0366. This suggests that generic subword tokenizers are ill-suited for geometric regression due to the resulting variable-length sequences—often exceeding 9 tokens—and the increased autoregressive burden. Unlike our training-free, uniform discretization strategy, the FastTokenizer incurs additional training overhead and redundancy without offering accuracy gains, confirming that our custom tokenizer provides a more stable and effective interface for geometric grounding.

\section{Additional Related Work}
\textbf{Assembly Datasets and Benchmarks.}
Robotic assembly datasets largely stem from object reassembly and part-based reconstruction, and can be broadly categorized into everyday object assembly, furniture assembly, fragment reassembly, and robotic assembly benchmarks. 
Representative datasets include Assembly101~\cite{sener2022assembly101} for multi-view toy-vehicle assembly, IKEAASM~\cite{ben2021ikea} for furniture assembly with fine-grained annotations, BreakingBad~\cite{sellan2022breaking} for large-scale fractured-object reconstruction, and ReAssemble~\cite{sliwowski2025reassemble} for narrow robotic tasks such as peg-in-hole and gear assembly. 
Recent studies further leverage VLMs to generate IKEA-style manuals~\cite{liu2024ikea,tie2025manual2skill,wang2022ikea,zhang2025manual}, but they remain mostly limited to furniture scenarios and image or point-cloud modalities. 
In contrast, we introduce AssemBench, a large-scale multimodal benchmark with over 900K samples and 150K assembly steps, enabling systematic research on multimodal robotic assembly at scale.

\section{Qualitative Visualizations and Dataset Examples}
To provide a more intuitive understanding of AssemLM's reasoning process and the richness of our training data, we present qualitative visualizations of both the dataset assets and the model's predictive performance.

As shown in Fig.~\ref{fig:furniture1} and Fig.~\ref{fig:daily_frag}, AssemBench provides high-fidelity multimodal supervision, covering a vast array of geometric structures and artistic rendering styles for instruction manuals. This diversity is key to the model's robust cross-modal alignment.

Furthermore, Fig.~\ref{fig:prediction} demonstrates AssemLM’s predictive accuracy on unseen test assets. The high degree of overlap between the predicted poses and the ground-truth configurations highlights the model's ability to resolve fine-grained spatial constraints. Even when faced with ambiguous manual instructions or sparse point clouds, the model generates physically plausible and precise assembly anchors, confirming the effectiveness of our pose discretization and multimodal fusion strategy.

\input{contents/fig_furniture1}
\input{contents/fig_daily_frag}
\input{contents/fig_prediction}

\end{appendix}

%% file: contents/fig_real_assets.tex
\begin{figure}[h!]
    \centering
    \includegraphics[width=1.0\textwidth]{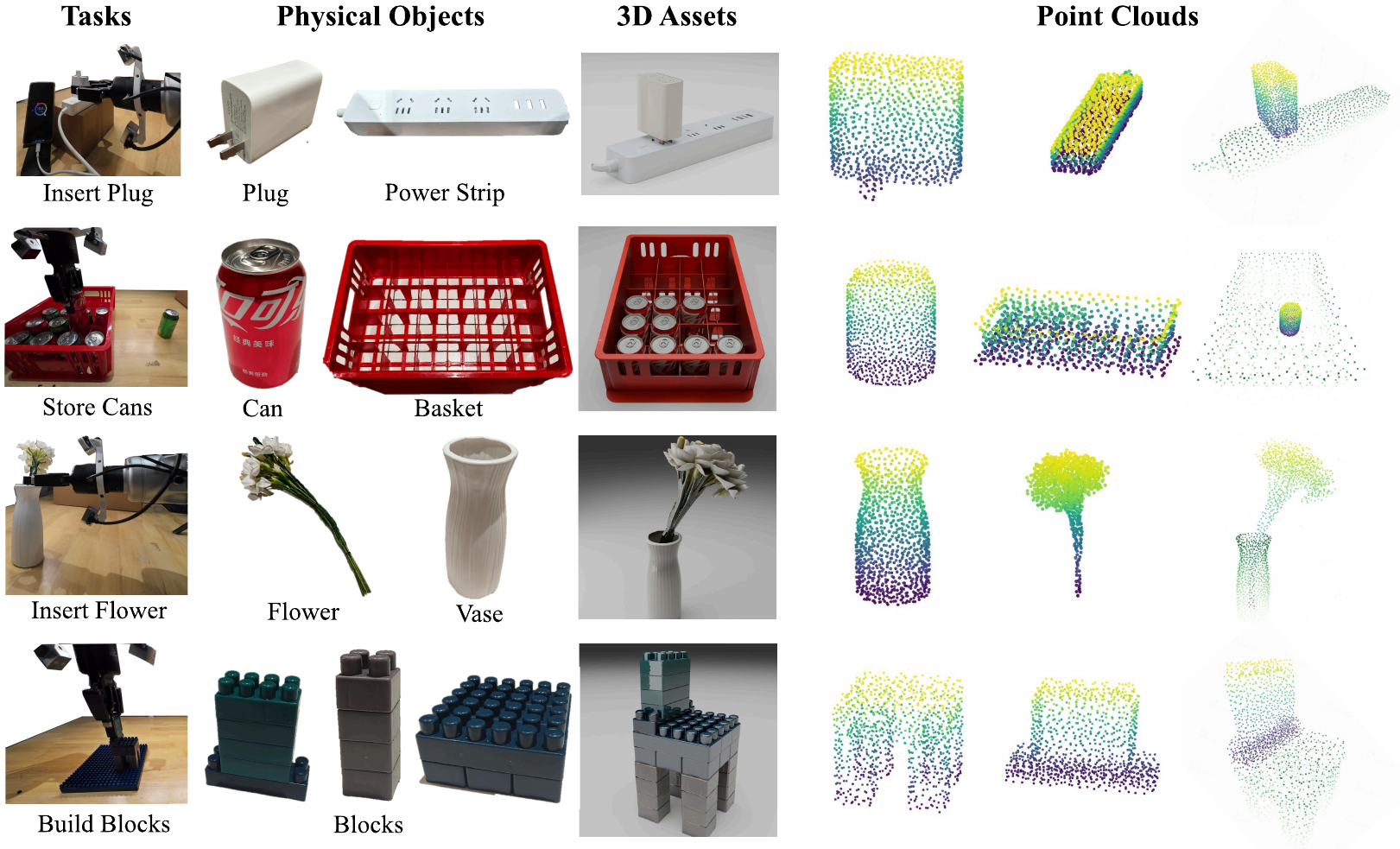}
    \caption{\textbf{Visualization of real-world asset processing.} We illustrate the data pipeline for four manipulation tasks: \textit{Insert Plug}, \textit{Store Cans}, \textit{Insert Flower}, and \textit{Build Blocks}. For each task, the figure displays the physical setup, the individual physical objects, the reconstructed high-fidelity 3D assets, and the final sampled point clouds used for model inference.}
    \label{fig:real_assts}
\end{figure}

%% file: contents/fig_canonical_coordinate_system.tex
\begin{wrapfigure}{r}{0.5\textwidth}
    \centering
    \vspace{-13pt}
    \includegraphics[width=1\linewidth]{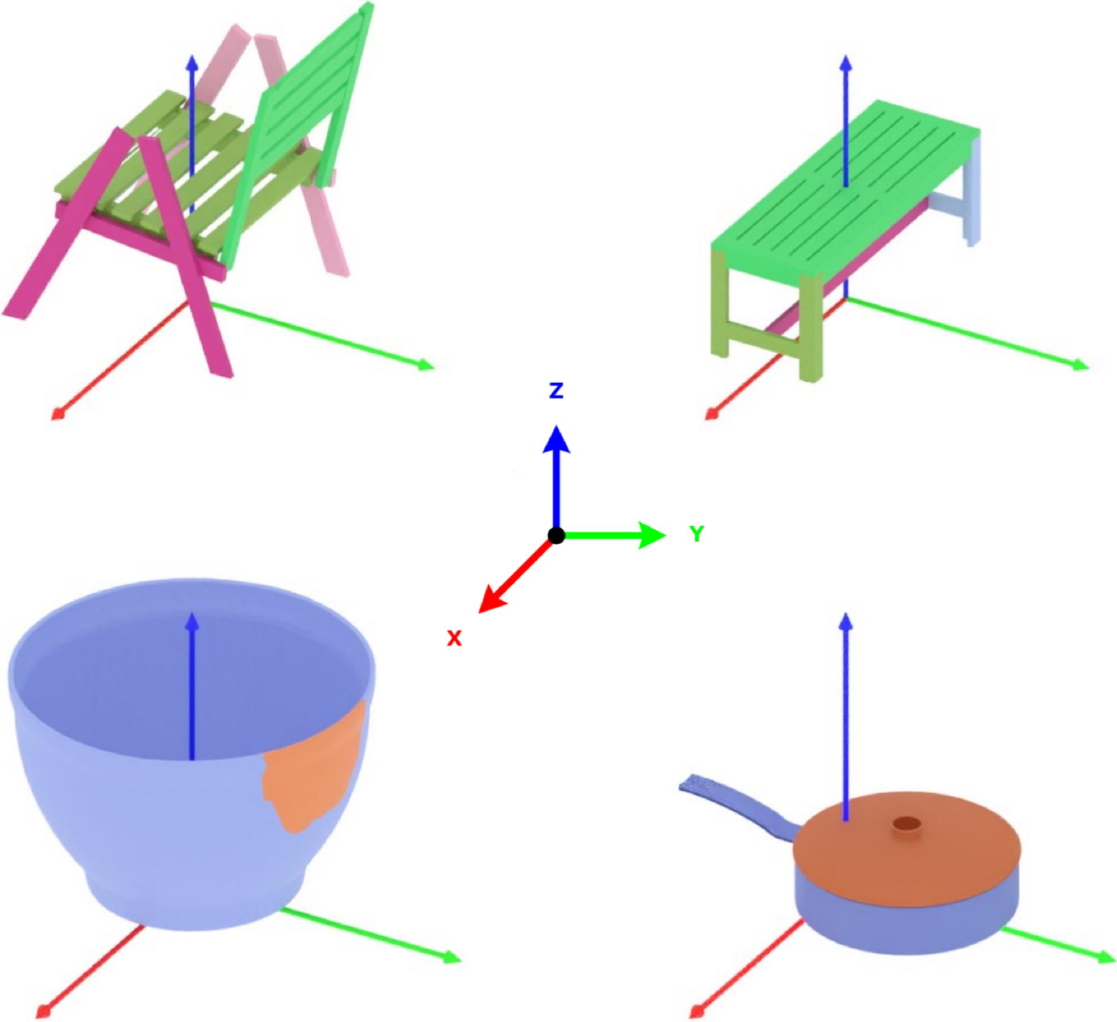}
    \vspace{-13pt}
    \caption{Canonical Coordinate System.}
    \vspace{-12pt}
    \label{fig:canonical_coordinate_system}
\end{wrapfigure}

%% file: contents/fig_real_step_and_manual.tex
\begin{figure*}[t]
    \centering
    \includegraphics[width=1\linewidth]{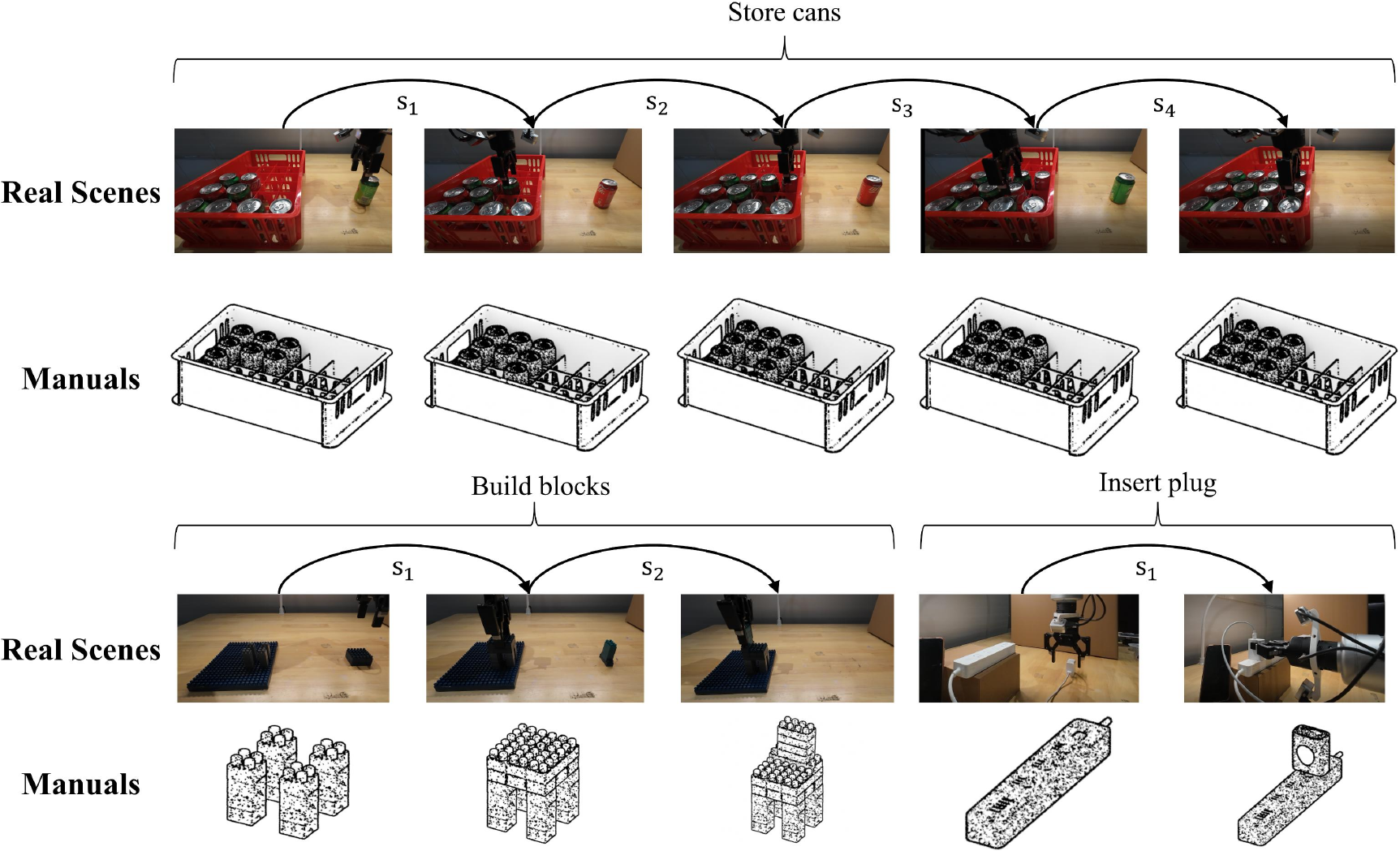}
    \caption{Real-world assembly steps and corresponding instruction manuals.}
    \label{fig:real_step_and_manual}
\end{figure*}

%% file: contents/table_ablition.tex
\begin{table*}[ht]
\centering
\caption{Ablation studies on model modalities and architectural modules.}
\resizebox{1\textwidth}{!}{
\begin{tabular}{l|cccc|cccc|cccc}
\toprule
\multirow{2}{*}{\textbf{}}
& \multicolumn{4}{c|}{\textbf{AssemLM w/o Manual}}
& \multicolumn{4}{c|}{\textbf{AssemLM w/o Instruction}}
& \multicolumn{4}{c}{\textbf{AssemLM (Ours)}} \\
& \textbf{R-T} \(\downarrow\) & \textbf{SCD(R)} \(\downarrow\)  & \textbf{SCD} \(\downarrow\) & \textbf{SR} \(\uparrow\)
& \textbf{R-T} \(\downarrow\) & \textbf{SCD(R)} \(\downarrow\)  & \textbf{SCD} \(\downarrow\) & \textbf{SR} \(\uparrow\)
& \textbf{R-T} \(\downarrow\) & \textbf{SCD(R)} \(\downarrow\)  & \textbf{SCD} \(\downarrow\) & \textbf{SR} \(\uparrow\) \\
\midrule

DO
& 0.0510 & 0.0066 & 0.0335 & 76.3\%
& 0.0381 & 0.0065 & 0.0135 & 82.6\%
& 0.0317 & 0.0060 & 0.0096 & 86.9\% \\

Fur.
& 0.1569 & 0.0119 & 0.2115 & 33.0\%
& 0.0249 & 0.0113 & 0.0233 & 74.0\%
& 0.0194 & 0.0029 & 0.0139 & 91.5\% \\

Frag.
& 0.0770 & 0.0081 & 0.1058 & 67.4\%
& 0.0127 & 0.0083 & 0.0142 & 86.4\%
& 0.0097 & 0.0062 & 0.0092 & 89.8\% \\

Fur.*
& 0.1853 & 0.0245 & 0.2429 & 14.8\%
& 0.0570 & 0.0206 & 0.0727 & 56.1\%
& 0.0557 & 0.0097 & 0.0613 & 69.6\% \\

\midrule

\multirow{2}{*}{\textbf{}}
& \multicolumn{4}{c|}{\textbf{AssemLM w/o DeepStack}}
& \multicolumn{4}{c|}{\textbf{AssemLM w/o Equivariant}}
& \multicolumn{4}{c}{\textbf{AssemLM w/o Warm-up Stage}} \\
& \textbf{R-T} \(\downarrow\) & \textbf{SCD(R)} \(\downarrow\)  & \textbf{SCD} \(\downarrow\) & \textbf{SR} \(\uparrow\)
& \textbf{R-T} \(\downarrow\) & \textbf{SCD(R)} \(\downarrow\)  & \textbf{SCD} \(\downarrow\) & \textbf{SR} \(\uparrow\)
& \textbf{R-T} \(\downarrow\) & \textbf{SCD(R)} \(\downarrow\)  & \textbf{SCD} \(\downarrow\) & \textbf{SR} \(\uparrow\) \\

\midrule
DO
& 0.0697 & 0.0067 & 0.0431 & 68.1\%
& 0.0379 & 0.0176 & 0.0258 & 61.8\%
& 0.0368 & 0.0061 & 0.0120 & 81.9\% \\

Fur.
& 0.0440 & 0.0094 & 0.0335 & 74.0\%
& 0.0215 & 0.0217 & 0.0339 & 63.5\%
& 0.0227 & 0.0124 & 0.0253 & 78.5\% \\

Frag.
& 0.0684 & 0.0082 & 0.0668 & 64.1\%
& 0.0096 & 0.0127 & 0.0166 & 77.6\%
& 0.0126 & 0.0088 & 0.0146 & 85.7\% \\

Fur.*
& 0.0741 & 0.0163 & 0.0755 & 49.8\%
& 0.0566 & 0.0354 & 0.0853 & 34.7\%
& 0.0578 & 0.0197 & 0.0707 & 55.8\% \\

\bottomrule
\end{tabular}
}
\vspace{-0.05in}
\label{tab:ablition}
\end{table*}

%% file: contents/table_addition_compare.tex
\begin{wraptable}{r}{0.5\textwidth}
\caption{Additional baseline methods.}
\centering
\resizebox{0.5\textwidth}{!}
{
\begin{tabular}{l|cccc|cccc}
\toprule
\multirow{2}{*}{\textbf{}}
& \multicolumn{4}{c|}{\textbf{ManualPA~\cite{zhang2025manual}}}
& \multicolumn{4}{c}{\textbf{SE(3)-Assembly~\cite{wu2023leveraging}}} \\
& \textbf{R-T} \(\downarrow\) & \textbf{SCD(R)} \(\downarrow\)  & \textbf{SCD} \(\downarrow\) & \textbf{SR} \(\uparrow\)
& \textbf{R-T} \(\downarrow\) & \textbf{SCD(R)} \(\downarrow\)  & \textbf{SCD} \(\downarrow\) & \textbf{SR} \(\uparrow\) \\

\midrule
DO
& 0.0787 & 0.0075 & 0.0122 & 83.4\%
& 0.1525 & 0.0186 & 0.0532 & 15.1\% \\

Fur.
& 0.1073 & 0.0071 & 0.0170 & 71.5\%
& 0.1583 & 0.0386 & 0.0975 & 10.5\% \\

Frag.
& 0.0918 & 0.0052 & 0.0101 & 89.1\%
& 0.16 & 0.0162 & 0.0497 & 11.3\% \\

Fur.*
& 0.1041 & 0.0129 & 0.0216 & 54.9\%
& 0.1551 & 0.0451 & 0.0839 & 2.1\% \\
\bottomrule
\end{tabular}
}

\label{tab:additional_baseline}
\end{wraptable}

%% file: contents/list_prompt.tex
\begin{lstlisting}[caption={The multimodal prompt template utilized for foundation model inference.}]
I will provide you with images showing the states before and after assembly, and their 3D point clouds. Your goal is to predict the transformation (9D pose) of the source part (Part A) to assemble it with the target part (Part B).

Source Point Cloud (Part A): (x1, y1, z1) ... (x1024, y1024, z1024)

Target Point Cloud (Part B): (x1, y1, z1) ... (x1024, y1024, z1024)

Please provide your answer as a 9D pose list: [trans_x, trans_y, trans_z, rot_r1, rot_r2, rot_r3, rot_r4, rot_r5, rot_r6]. 

- [trans_x, trans_y, trans_z] represents the 3D translation vector.

- [rot_r1, rot_r2, rot_r3] and [rot_r4, rot_r5, rot_r6] represent the first and second columns of the continuous 6D rotation matrix representation respectively.

Only output the list.

[Visual Inputs Group]:
<Manual_1> <Manual_2>
\end{lstlisting}

%% file: contents/table_exp_1_full.tex
\vspace{-0.03in}
\begin{table}[ht!] 
\centering

\caption{\textbf{Quantitative Results across Multi-category Assembly Tasks.} We evaluate AssemLM against a specialized $SE(3)$-equivariant model (TwoByTwo) and state-of-the-art foundation models (GPT-5.2 and DeepSeek-V3.2). Performance is reported via Translation RMSE (RMSE(T)), Symmetric Chamfer Distance (SCD), and Success Rate (SR, defined as $SCD < 0.02$). For cross-dataset consistency, related sub-categories (e.g., Bottle variants) are consolidated. The "All" row reflects the weighted average across Daily Objects, Furniture, and Fragments. Bold indicates the best performance in each category.}
\resizebox{1.0\textwidth}{!}{
\begin{tabular}{l|ccc|ccc|ccc|ccc}
\toprule
\multirow{2}{*}{\textbf{Category}} 
& \multicolumn{3}{c|}{\textbf{TwoByTwo~\cite{qi2025two}}} 
& \multicolumn{3}{c|}{\textbf{GPT-5.2~\cite{singh2025openai}}} 
& \multicolumn{3}{c|}{\textbf{DeepSeek-V3.2~\cite{liu2025deepseek}}} 
& \multicolumn{3}{c}{\textbf{AssemLM (Ours)}} \\
& \textbf{RMSE(T)} $\downarrow$ & \textbf{SCD} $\downarrow$ & \textbf{SR} $\uparrow$
& \textbf{RMSE(T)} $\downarrow$ & \textbf{SCD} $\downarrow$ & \textbf{SR} $\uparrow$
& \textbf{RMSE(T)} $\downarrow$ & \textbf{SCD} $\downarrow$ & \textbf{SR} $\uparrow$
& \textbf{RMSE(T)} $\downarrow$ & \textbf{SCD} $\downarrow$ & \textbf{SR} $\uparrow$ \\
\midrule

KitchenPort     
& 0.1544 & 0.1131 & 16.7\%
& 0.1224 & 0.0896 & 0.0\%
& 0.2181 & 0.2252 & 0.0\%
& \textbf{0.0374} & \textbf{0.0153} & \textbf{71.4\%} \\

Bottle
& 0.1953 & 0.1692 & 16.7\%
& 0.2052 & 0.1728 & 19.2\%
& 0.1826 & 0.2564 & 0.0\%
& \textbf{0.0226} & \textbf{0.0102} & \textbf{88.9\%} \\

Kettle          
& 0.1148 & 0.0631 & 20.0\%
& 0.2066 & 0.1665 & 0.0\%
& 0.2773 & 0.3789 & 0.0\%
& \textbf{0.0306} & \textbf{0.0086} & \textbf{66.7\%} \\

Coffeemachine   
& 0.2833 & 0.3311 & 0.0\%
& 0.2231 & 0.1525 & 0.0\%
& 0.3412 & 0.3044 & 0.0\%
& \textbf{0.0304} & \textbf{0.0051} & \textbf{100.0\%} \\

Cup             
& 0.1583 & 0.0777 & 0.0\%
& 0.1318 & 0.0856 & 0.0\%
& 0.1792 & 0.2585 & 0.0\%
& \textbf{0.0066} & \textbf{0.0014} & \textbf{100.0\%} \\

Plug            
& 0.0483 & 0.0285 & 66.7\%
& 0.2833 & 0.3340 & 0.0\%
& 0.3396 & 0.6262 & 0.0\%
& \textbf{0.0340} & \textbf{0.0053} & \textbf{100.0\%} \\

Childrentoy     
& 0.1135 & 0.0737 & 50.0\%
& 0.2884 & 0.3269 & 6.2\%
& 0.3576 & 0.5727 & 0.0\%
& \textbf{0.0763} & \textbf{0.0271} & \textbf{77.8\%} \\

Postbox         
& 0.2710 & 0.3005 & 0.0\%
& 0.1908 & 0.1823 & 0.0\%
& 0.3050 & 0.5379 & 0.0\%
& \textbf{0.0373} & \textbf{0.0170} & \textbf{78.6\%} \\

Toaster         
& 0.0899 & 0.0363 & 14.3\%
& 0.3051 & 0.2849 & 0.0\%
& 0.2828 & 0.2989 & 0.0\%
& \textbf{0.0619} & \textbf{0.0138} & \textbf{57.1\%} \\

Nut             
& 0.0841 & 0.0879 & 12.5\%
& 0.1519 & 0.0830 & 41.7\%
& 0.1964 & 0.2006 & 16.7\%
& \textbf{0.0032} & \textbf{0.0035} & \textbf{100.0\%} \\

Coin            
& 0.1366 & 0.0975 & 50.0\%
& 0.1925 & 0.2541 & 12.5\%
& 0.2500 & 0.4180 & 0.0\%
& \textbf{0.0417} & \textbf{0.0109} & \textbf{87.5\%} \\

Key             
& 0.2624 & 0.3331 & 0.0\%
& 0.1882 & 0.1597 & 0.0\%
& 0.2503 & 0.2695 & 0.0\%
& \textbf{0.0095} & \textbf{0.0038} & \textbf{90.0\%} \\

Usb             
& 0.2496 & 0.2119 & 0.0\%
& 0.3560 & 0.5209 & 0.0\%
& 0.3415 & 0.5744 & 0.0\%
& \textbf{0.0216} & \textbf{0.0050} & \textbf{87.5\%} \\

Plate           
& 0.0509 & 0.0181 & 63.6\%
& 0.1553 & 0.1067 & 18.2\%
& 0.2468 & 0.3159 & 0.0\%
& \textbf{0.0298} & \textbf{0.0075} & \textbf{100.0\%} \\

Flower          
& 0.0954 & 0.0535 & 20.0\%
& 0.2246 & 0.2166 & 0.0\%
& 0.3077 & 0.4961 & 0.0\%
& \textbf{0.0363} & \textbf{0.0081} & \textbf{88.2\%} \\

Toilet          
& 0.0990 & 0.0223 & 62.5\%
& 0.2687 & 0.1964 & 12.5\%
& 0.3346 & 0.4260 & 0.0\%
& \textbf{0.0260} & \textbf{0.0050} & \textbf{100.0\%} \\

\midrule
\textbf{Daily Obj.} 
& 0.1504 & 0.1266 & 24.3\%
& 0.1923 & 0.1788 & 17.2\%
& 0.2633 & 0.3350 & 3.6\%
& \textbf{0.0317} & \textbf{0.0096} & \textbf{86.9\%} \\
\midrule

Chair              
& 0.1665 & 0.1391 & 3.6\%
& 0.2718 & 0.3260 & 1.9\%
& 0.2938 & 0.4427 & 1.2\%
& \textbf{0.0161} & \textbf{0.0076} & \textbf{95.6\%} \\

Table              
& 0.2023 & 0.2213 & 1.4\%
& 0.2278 & 0.2904 & 1.3\%
& 0.3262 & 0.6720 & 3.7\%
& \textbf{0.0245} & \textbf{0.0227} & \textbf{82.5\%} \\

Storage
& 0.1400 & 0.1340 & 14.3\%
& 0.1362 & 0.1880 & 5.4\%
& 0.2451 & 0.4409 & 3.6\%
& \textbf{0.0215} & \textbf{0.0205} & \textbf{95.8\%} \\

\midrule
\textbf{Furniture} 
& 0.1761 & 0.1669 & 4.0\%
& 0.2119 & 0.2681 & 2.9\%
& 0.2884 & 0.5185 & 2.8\%
& \textbf{0.0194} & \textbf{0.0139} & \textbf{91.5\%} \\
\midrule

Mirror            
& 0.1982 & 0.1856 & 4.9\%
& 0.2660 & 0.3027 & 4.7\%
& 0.2779 & 0.3818 & 4.3\%
& \textbf{0.0152} & \textbf{0.0179} & \textbf{91.3\%} \\

Plate
& 0.2210 & 0.2089 & 1.8\%
& 0.2747 & 0.3115 & 2.3\%
& 0.2820 & 0.4058 & 2.1\%
& \textbf{0.0120} & \textbf{0.0088} & \textbf{89.7\%} \\

Toy Figure        
& 0.2041 & 0.2005 & 5.6\%
& 0.2748 & 0.3142 & 2.2\%
& 0.2777 & 0.4153 & 2.1\%
& \textbf{0.0123} & \textbf{0.0094} & \textbf{91.1\%} \\

Mug               
& 0.1693 & 0.1414 & 7.3\%
& 0.2664 & 0.3017 & 4.2\%
& 0.2777 & 0.3966 & 3.8\%
& \textbf{0.0115} & \textbf{0.0133} & \textbf{89.0\%} \\

Bottle (Merged)            
& 0.1403 & 0.1102 & 34.7\%
& 0.2682 & 0.2985 & 4.2\%
& 0.2754 & 0.3889 & 3.9\%
& \textbf{0.0043} & \textbf{0.0038} & \textbf{98.3\%} \\

Bowl              
& 0.1922 & 0.1668 & 4.3\%
& 0.2666 & 0.3064 & 4.1\%
& 0.2772 & 0.3980 & 3.7\%
& \textbf{0.0102} & \textbf{0.0099} & \textbf{90.5\%} \\

Utensil
& 0.1642 & 0.1054 & 9.4\%
& 0.2661 & 0.3038 & 4.2\%
& 0.2774 & 0.3977 & 3.8\%
& \textbf{0.0043} & \textbf{0.0064} & \textbf{97.0\%} \\

Vase              
& 0.1983 & 0.1724 & 13.8\%
& 0.2664 & 0.3058 & 4.0\%
& 0.2767 & 0.3972 & 3.7\%
& \textbf{0.0151} & \textbf{0.0185} & \textbf{87.4\%} \\

Wine Glass        
& 0.2036 & 0.1551 & 5.9\%
& 0.2662 & 0.3052 & 4.1\%
& 0.2773 & 0.3972 & 3.7\%
& \textbf{0.0033} & \textbf{0.0074} & \textbf{81.8\%} \\

Teapot            
& 0.1997 & 0.1958 & 7.5\%
& 0.2662 & 0.3060 & 4.2\%
& 0.2775 & 0.3975 & 3.8\%
& \textbf{0.0057} & \textbf{0.0026} & \textbf{95.2\%} \\

Teacup
& 0.1695 & 0.1233 & 6.3\%
& 0.2661 & 0.3051 & 4.3\%
& 0.2775 & 0.3975 & 3.9\%
& \textbf{0.0048} & \textbf{0.0077} & \textbf{78.6\%} \\

Statue
& 0.1400 & 0.0645 & 25.0\%
& 0.2666 & 0.3058 & 4.0\%
& 0.2773 & 0.3978 & 3.6\%
& \textbf{0.0093} & \textbf{0.0064} & \textbf{100.0\%} \\

Ring
& 0.1481 & 0.1050 & 0.0\%
& 0.2664 & 0.3058 & 4.0\%
& 0.2773 & 0.3975 & 3.6\%
& \textbf{0.0220} & \textbf{0.0185} & \textbf{66.7\%} \\

Cookie
& 0.2188 & 0.1920 & 0.0\%
& 0.2660 & 0.3055 & 4.0\%
& 0.2770 & 0.3977 & 3.6\%
& \textbf{0.0110} & \textbf{0.0078} & \textbf{90.0\%} \\

Spoon
& 0.0443 & 0.0031 & \textbf{100.0\%}
& 0.3775 & 0.5806 & 0.0\%
& 0.2741 & 0.2774 & 0.0\%
& \textbf{0.0050} & \textbf{0.0003} & \textbf{100.0\%} \\

\midrule
\textbf{Fragments}
& 0.1741 & 0.1420 & 15.1\%
& 0.2670 & 0.3067 & 3.8\%
& 0.2724 & 0.3950 & 3.4\%
& \textbf{0.0097} & \textbf{0.0092} & \textbf{89.8\%} \\
\midrule

\textbf{All}
& 0.1669 & 0.1452 & 14.5\%
& 0.2238 & 0.2512 & 7.9\%
& 0.2747 & 0.4162 & 3.3\%
& \textbf{0.0203} & \textbf{0.0109} & \textbf{89.4\%} \\

\bottomrule
\end{tabular}
}
\vspace{-0.15in}
\label{tab:main_results_multi_category_full}
\end{table}

%% file: contents/fig_ablition.tex
\begin{figure*}
    \centering
    \includegraphics[width=1\linewidth]{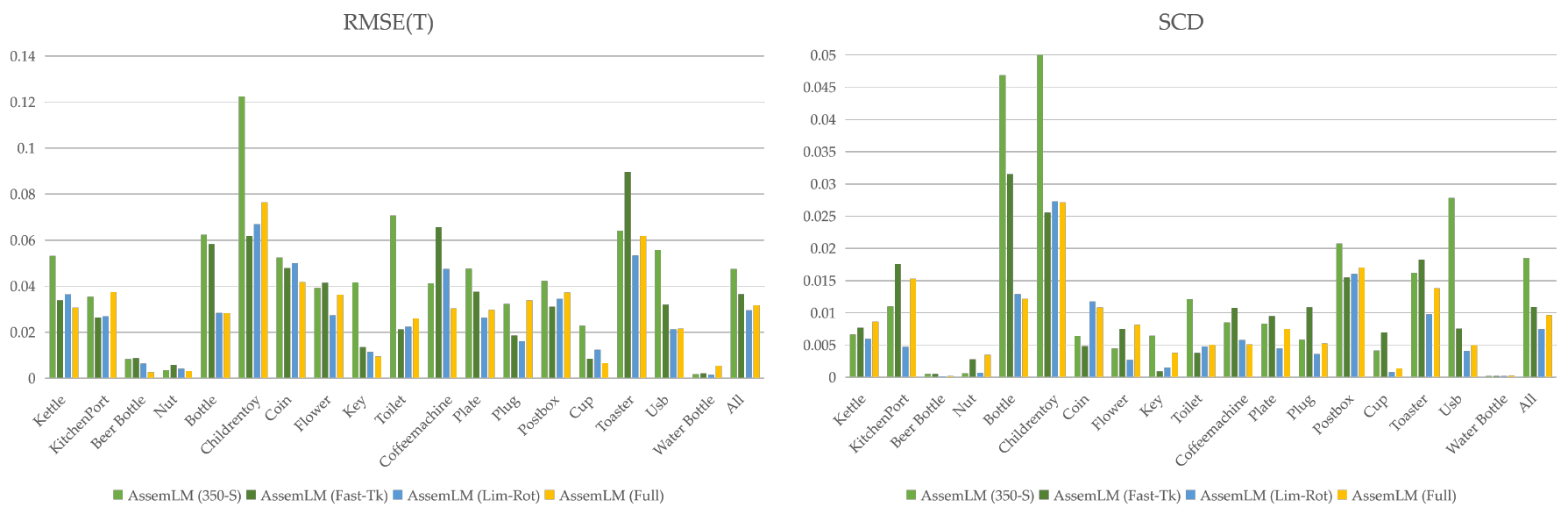}
\caption{\textbf{Ablation study on dataset scale, rotation range, and tokenizer choice.} 
    We evaluate the impact of different design components using Translation RMSE (left) and Symmetric Chamfer Distance (right) across diverse daily objects. 
    The \textbf{All} bars represent the average across all categories.}
    \label{fig:ablation_analysis}
\end{figure*}

%% file: contents/fig_furniture1.tex
\begin{figure*}[t]
    \centering
    \includegraphics[width=1\linewidth]{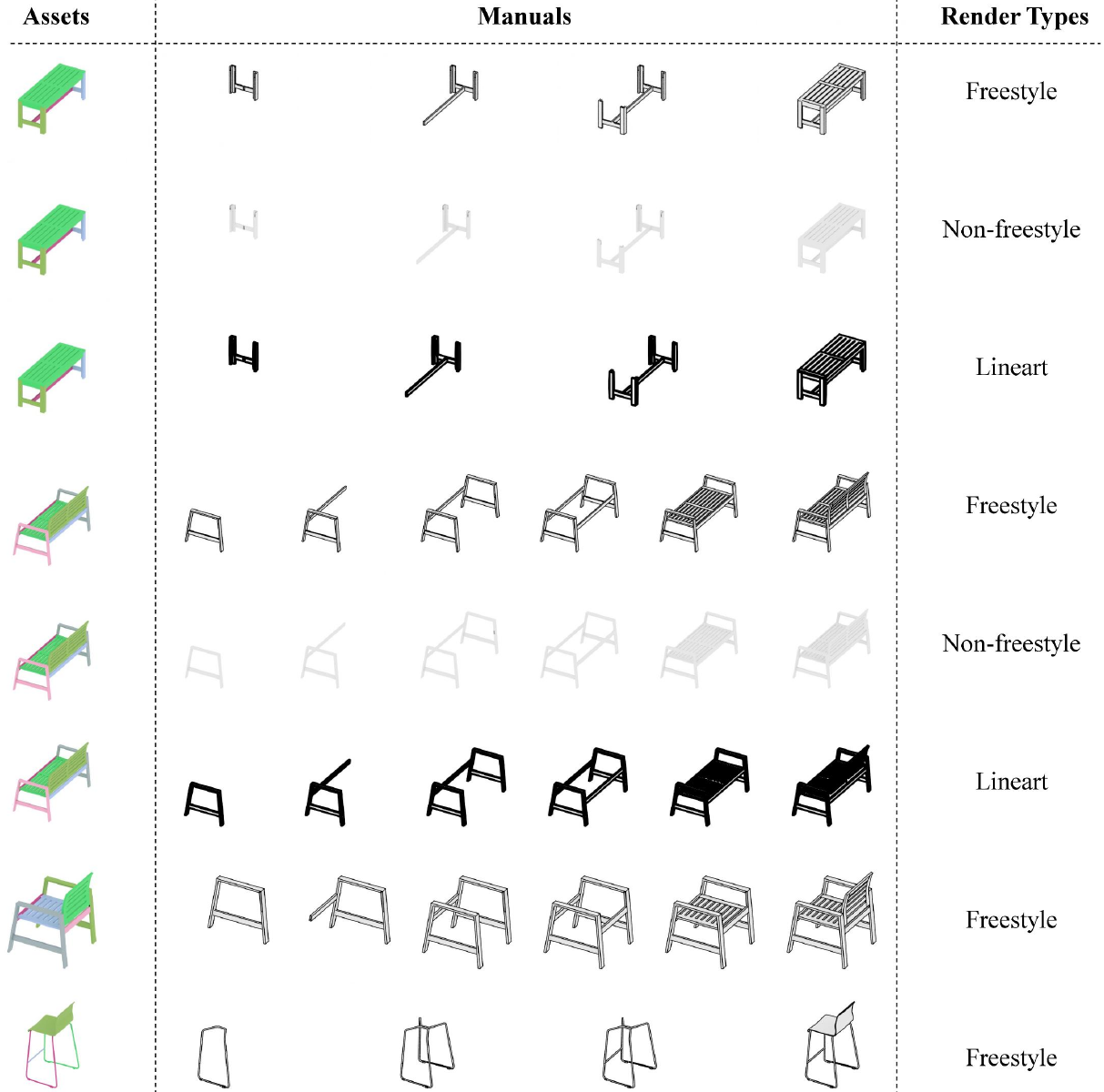}
    \caption{Visualization of Different Types of Manuals.}
    \label{fig:furniture1}
\end{figure*}

%% file: contents/fig_daily_frag.tex
\begin{figure*}[t]
    \centering
    \includegraphics[width=1\linewidth]{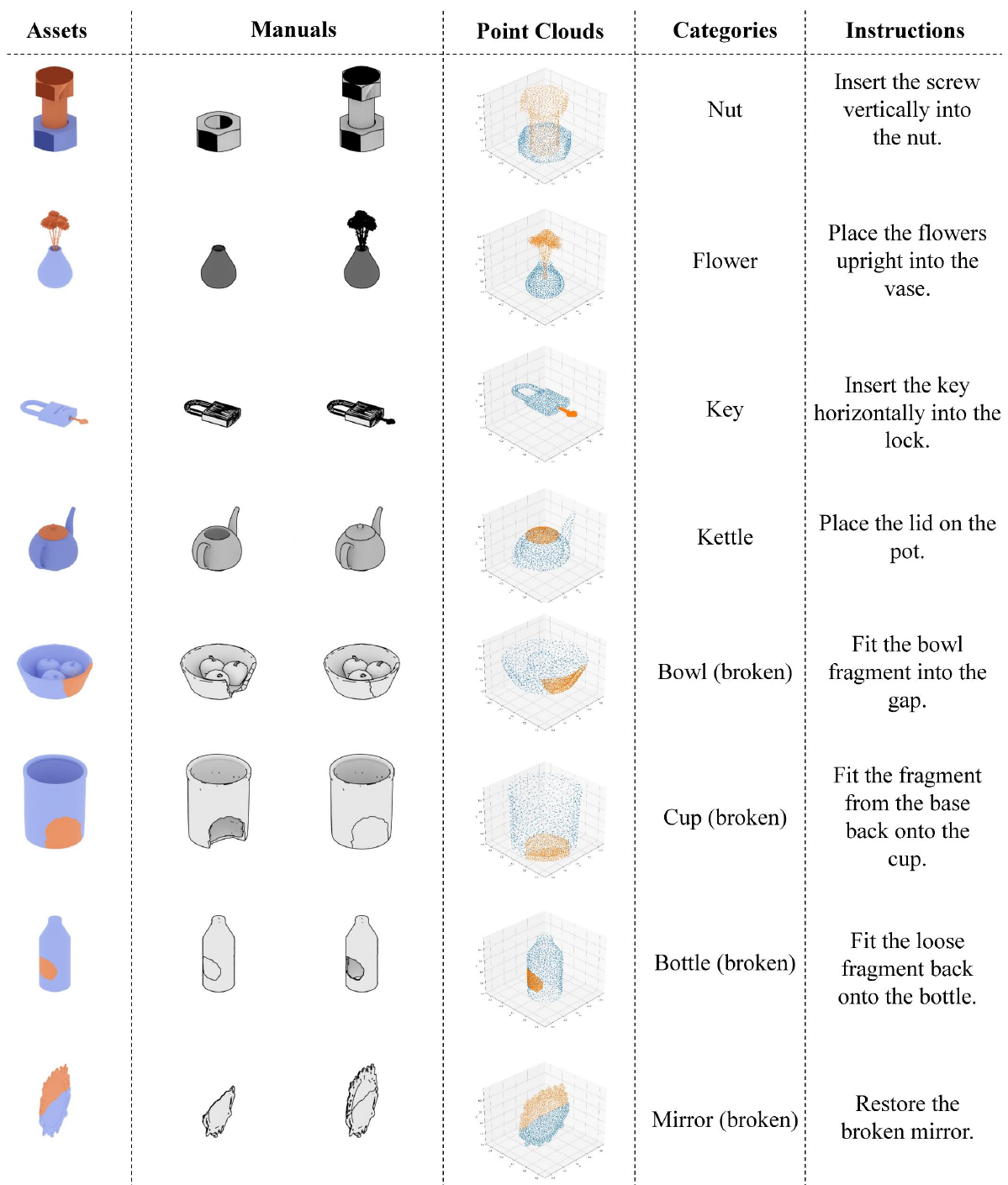}
    \caption{Representative examples from AssemBench.}
    \label{fig:daily_frag}
\end{figure*}

%% file: contents/fig_prediction.tex
\begin{figure*}[t]
    \centering
    \includegraphics[width=1\linewidth]{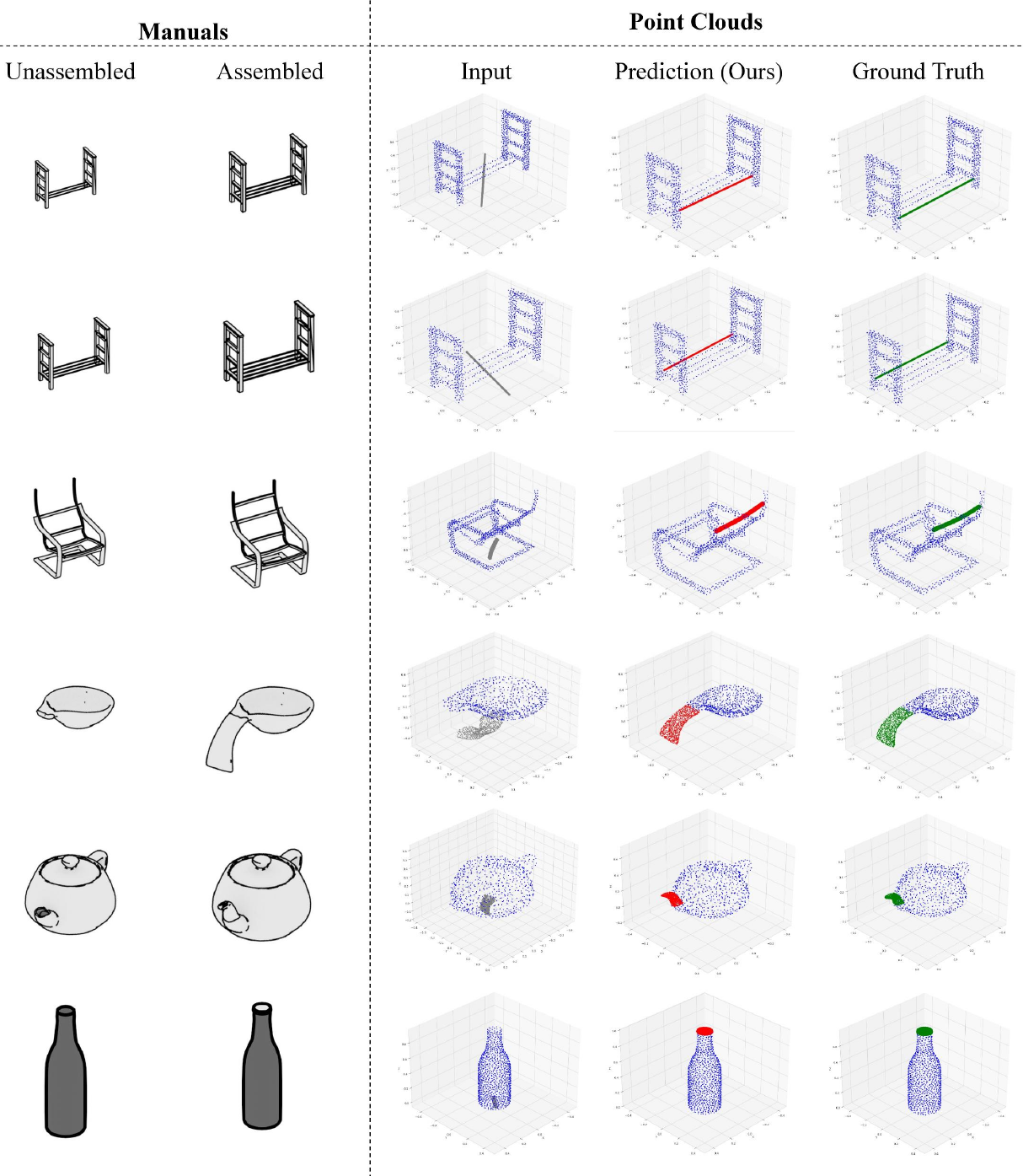}
    \caption{Visualization of assembly pose predictions by AssemLM.}
    \label{fig:prediction}
\end{figure*}